\documentclass[a4paper, 12pt]{article}

\usepackage[sort&compress]{natbib}
\bibpunct{(}{)}{;}{a}{}{,} 

\usepackage{amsthm, amsmath, amssymb, mathrsfs, multirow, url, subfigure}
\usepackage{float}
\usepackage{bm}
\usepackage{graphicx} 
\usepackage{verbatim}
\usepackage{algorithm2e}
\usepackage{natbib}
\usepackage{ifthen} 
\usepackage{amsfonts}
\usepackage[usenames]{color}
\usepackage{fullpage}
\usepackage{caption}

\theoremstyle{plain}

\theoremstyle{definition}

\theoremstyle{remark}
\newtheorem{remark}{Remark}

\newcommand{\RR}{\mathbb{R}}

\newcommand{\E}{\mathsf{E}}

\newcommand{\Qset}{\mathscr{Q}}

\newcommand{\eps}{\varepsilon}
\renewcommand{\phi}{\varphi}

\newcommand{\grad}{\nabla}

\newcommand{\nm}{\mathsf{N}}
\newcommand{\mvn}{\mathsf{MVN}}

\newcommand{\ber}{\mathsf{Ber}}

\title{Variational approximations using Fisher divergence}
\author{Yue Yang,\footnote{Department of Statistics, North Carolina State University; {\tt yyang44@ncsu.edu}, {\tt rgmarti3@ncsu.edu}} \quad Ryan Martin,$^*$ \quad Howard Bondell\footnote{{School of Mathematics and Statistics}, University of Melbourne; {\tt howard.bondell@unimelb.edu.au}}}
\date{\today}

\begin{document}

\maketitle 


\begin{abstract}
Modern applications of Bayesian inference involve models that are sufficiently complex that the corresponding posterior distributions are intractable and must be approximated. The most common approximation is based on Markov chain Monte Carlo, but these can be expensive when the data set is large and/or the model is complex, so more efficient variational approximations have recently received considerable attention. The traditional variational methods, that seek to minimize the Kullback--Leibler divergence between the posterior and a relatively simple parametric family, provide accurate and efficient estimation of the posterior mean, but often does not capture other moments, and have limitations in terms of the models to which they can be applied. Here we propose the construction of variational approximations based on minimizing the Fisher divergence, and develop an efficient computational algorithm that can be applied to a wide range of models without conjugacy or potentially unrealistic mean-field assumptions. We demonstrate the superior performance of the proposed method for the benchmark case of logistic regression.


\smallskip

\emph{Keywords and phrases:} Bayesian inference; exponential family; iteratively re-weighted least squares; logistic regression; Markov chain Monte Carlo.

\end{abstract}

\section{Introduction}
\label{S:intro}

Bayesian inference has become increasingly popular in the last 20+ years thanks to fundamental developments in Markov chain Monte Carlo methodology, providing computational tools for accurately approximating complex posterior distributions.  But despite the power of these methods, there are situations where the necessary computations are prohibitively slow or expensive, so other more efficient approximations may be desired.  In recent years, interest in so-called {\em variational approximations} has surged, thanks in part to their computational simplicity \citep[e.g.,][]{blei.etal.vb.2017}.  Roughly, this variational approach proceeds by identifying a sufficiently rich yet analytically tractable class of distributions and then chooses the member of that class which is closest to the posterior distribution with respect to some discrepancy measure.  The computational efficiency gain is a result of converting a difficult integration problem into an optimization problem for which there are many fast and powerful algorithms available.  

To set the scene, we first introduce some notation.  Since what we have in mind here are applications in Bayesian statistics, we will use notation consistent with that; however, the proposed approximation strategy can be applied more generally.  Suppose the quantity of interest is $\theta$, taking values in a space $\Theta \subseteq \RR^d$, and we are given both a prior density $\pi(\theta)$ and a likelihood function $L_y(\theta)$, the latter depending on data $y$.  Let $\pi_y$ denote the corresponding posterior density, and write $\tilde\pi_y(\theta) = L_y(\theta) \pi(\theta)$ for its unnormalized version.  The posterior density $\pi_y$ and its relevant features might be difficult to compute for one reason or another, e.g., the normalizing constant might not be available in closed-form.  Therefore, we seek a simpler and more tractable approximation to $\pi_y$.  A variational approach proceeds by introducing a family $\Qset$ of densities on $\Theta$ and a discrepancy measure $D$, and takes as the approximation a density $q^\star$ in $\Qset$ that minimizes $D(q, \pi_y)$.  

Implementation of the variational approach described above requires specification of a class of candidate approximations and a measure of discrepancy between densities.  Our focus in this paper is on the latter issue, namely, the choice of discrepancy measure.  In the variational approximation literature, the standard choice is the Kullback--Leibler divergence \citep{kullback.leibler.1951, kullback.book}, which has a number of desirable properties in this context.  Minimizing the Kullback--Leibler divergence of the posterior from the approximating family will emphasize certain features of the approximation, whereas other divergence measures will focus on other features.  Therefore, it is of interest to consider alternative discrepancy measures and what impact these might have on the quality of approximation.  For example, investigations into the use of R\'enyi or $\alpha$-divergences \citep{vanerven.renyi} have resulted in tighter marginal likelihood bounds and improved performance in certain applications \citep[e.g.,][]{li.turner.alpha, blei.etal.vb.2017}.  One challenge in dealing with complex distributions is that the normalizing constants are rarely available in closed form, so it would be advantageous if the discrepancy measure to be minimized did not depend on these normalizing constants.  One such discrepancy measure is the {\em Fisher divergence}, or Fisher information distance (\citealt{johnson.2004.book}, Definition~1.13; \citealt{dasgupta}, Definition~2.5), and here we investigate the performance of approximations based on minimizing the Fisher divergence.  

We start in Section~\ref{S:fisher} by defining the Fisher divergence, reviewing its basic properties, and giving some simple illustrations.  A key insight is that the Fisher divergence reduces to a relatively simple form when the approximating densities belong to an exponential family.  We exploit this simplified form in Section~\ref{S:approx} where we propose a fast iteratively re-weighted least squares algorithm to carry out the minimization, without requiring conjugacy in the Bayesian model or restrictive independent/mean-field approximations.  A practically important application is logistic regression, which serves as a benchmark test case for variational methods, and in Section~\ref{S:results} we demonstrate the superior performance of our Fisher divergence-driven approach compared to existing methods in terms of accuracy of the posterior approximation.  Concluding remarks are given in Section~\ref{S:discuss}.

\section{Fisher divergence}
\label{S:fisher}

When it comes to measuring the discrepancy between two density functions, the Kullback--Leibler divergence, Hellinger distance, and other kinds of  $f$-divergences \citep[e.g.,][]{liesevadja}, are the most widely used.  A somewhat less common discrepancy measure, however, is the {\em Fisher divergence}, defined as 
\begin{equation}
\label{eq:fisher}
F(q,p)=\int_{\RR^d} \|\grad \log q(\theta) - \grad \log p(\theta)\|^2 \, q(\theta) \,d\theta, 
\end{equation}
where $p$ and $q$ are suitably smooth density functions, $\grad$ is the gradient operator, which in this case returns a column $d$-vector, and $\|\cdot\|$ is the usual $\ell_2$-norm on $\RR^d$.  The Fisher divergence has proved useful in a variety of statistics and machine learning applications; see, for example, \citet{holmes.walker.scaling}, \citet{walker.2016.info}, \citet{huggins.etal.fisher}.  It has also been shown in \citet{johnson2004fisher} to have a fundamental connection to classical central limit theorems.  

One can immediately see that $F$ is not symmetric, so it is not a proper metric.  However, it has other desirable properties of a discrepancy measure, i.e., non-negative and equal zero if and only if the two densities are equal $Q$-almost everywhere, where $Q(d\theta) = q(\theta)\,d\theta$.  Moreover, it is easy to see that $F(q,p)$ does not depend on the normalizing constant associated with the density $p$.  Connections between Fisher divergence and certain ``rates of change'' in Kullback--Leibler divergence can be seen in de Bruijn's identity \citep[e.g.,][]{stam1959some, barron1986entropy} and Stein's identity \citep[e.g.,][]{liu2016stein}; see, also, \citep{park2012equivalence}.  

\begin{remark}
\label{re:huggins}
Connections between the Fisher divergence and Wasserstein distance were discussed recently in \citet{huggins.etal.fisher}.  They showed posterior approximations based on minimizing the Wasserstein distance, as opposed to Kullback--Leibler divergence, provides better error control on moments.  They also showed that a suitable Fisher divergence upper-bounds Wasserstein distance, which suggests that approximations based on minimizing the former, compared to Kullback--Leibler divergence, would lead to improved moment estimates.  However, their inequalities involve constants that depend on the distribution being approximated, so the relative moment estimation accuracy varies from case to case.  Specific examples that demonstrate this variability are given below; see, also, the logistic regression application in Section~\ref{S:results}.
\end{remark}

\begin{figure}[t]
\begin{center}
\subfigure[$\nu=1$]{\label{fig:t1}\includegraphics[width=0.3\textwidth]{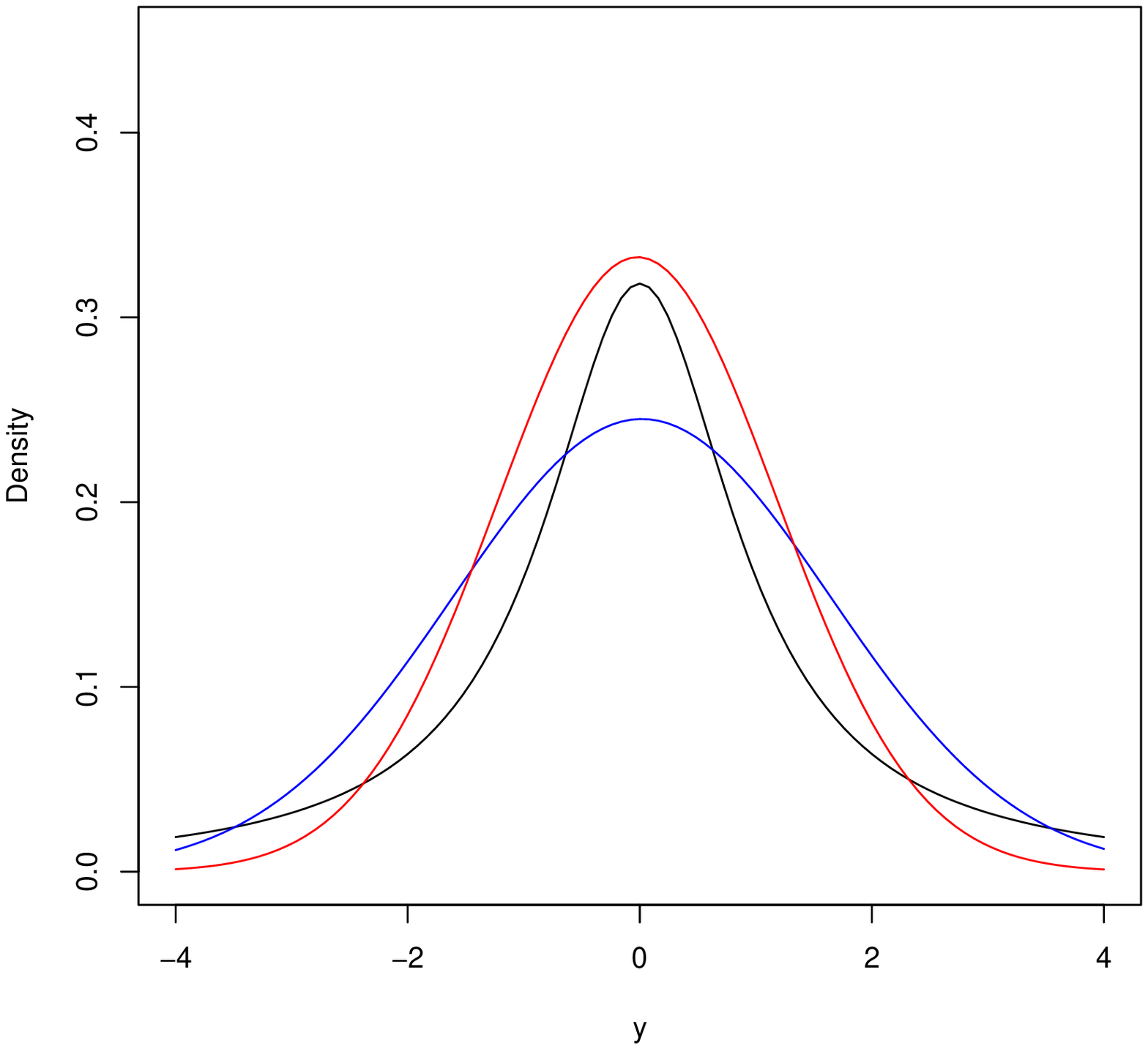}}
\subfigure[$\nu=2$]{\label{fig:t2}\includegraphics[width=0.3\textwidth]{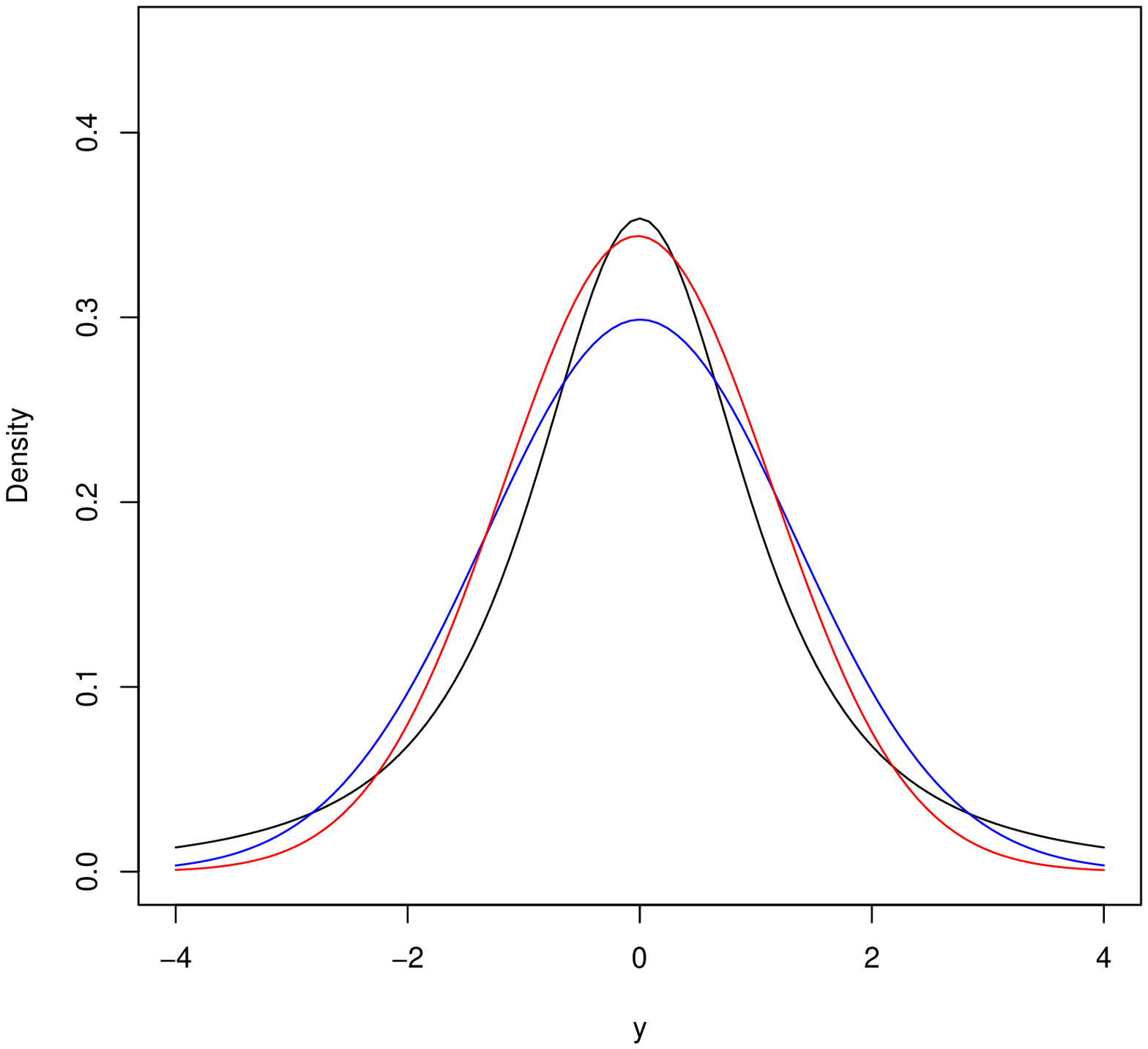}}
\subfigure[$\nu=4$]{\label{fig:t3}\includegraphics[width=0.3\textwidth]{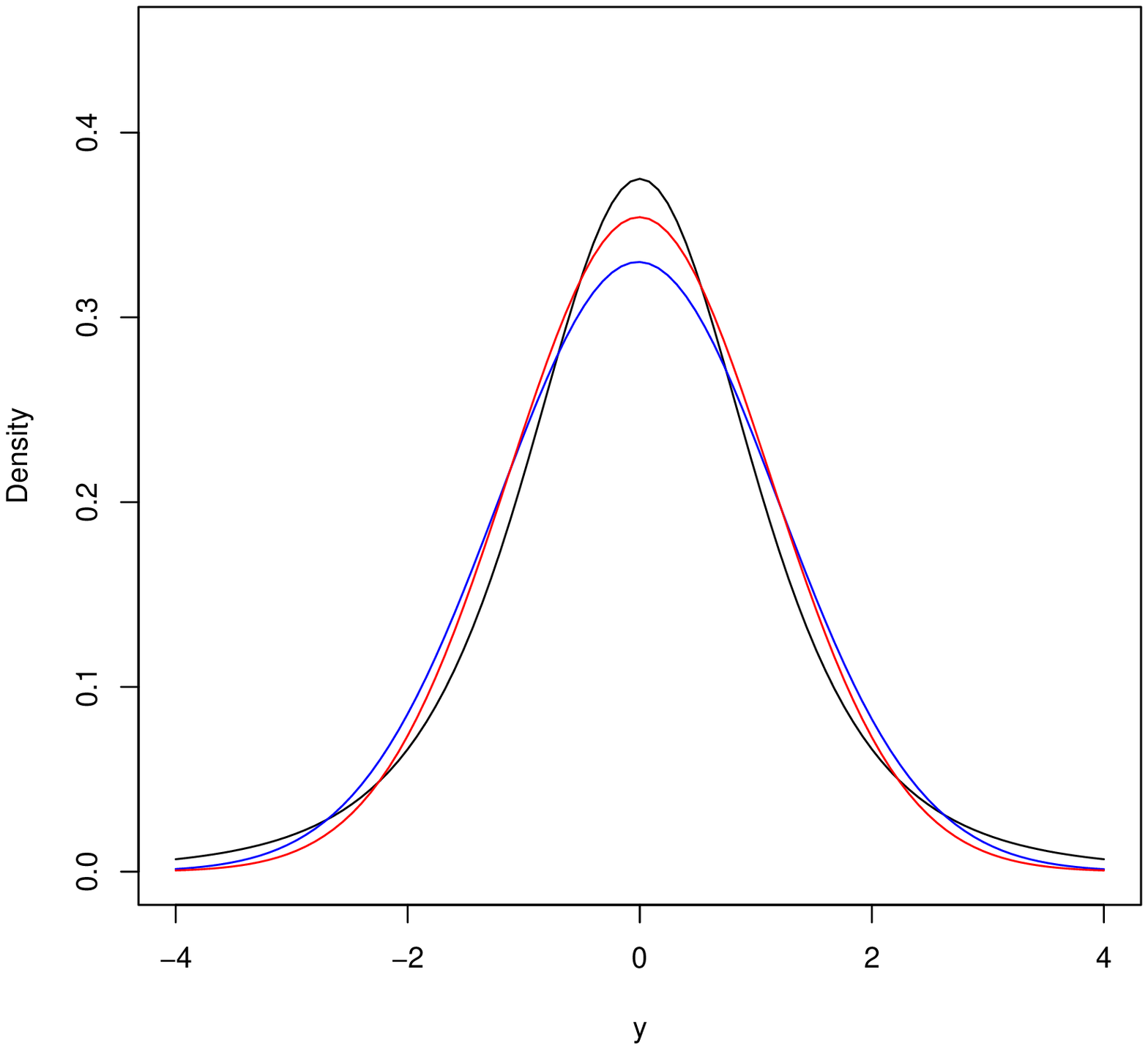}}
\end{center}
\caption{Approximation of a Student-t density (black) with degrees of freedom $\nu$ by a normal based on minimizing Kullback--Leibler (blue) and Fisher divergence (red).}
\label{fig:t}
\end{figure}
\begin{figure}[t]
\begin{center}
\subfigure[$\mu=1.5$]{\label{fig:mix1}\includegraphics[width=0.3\textwidth]{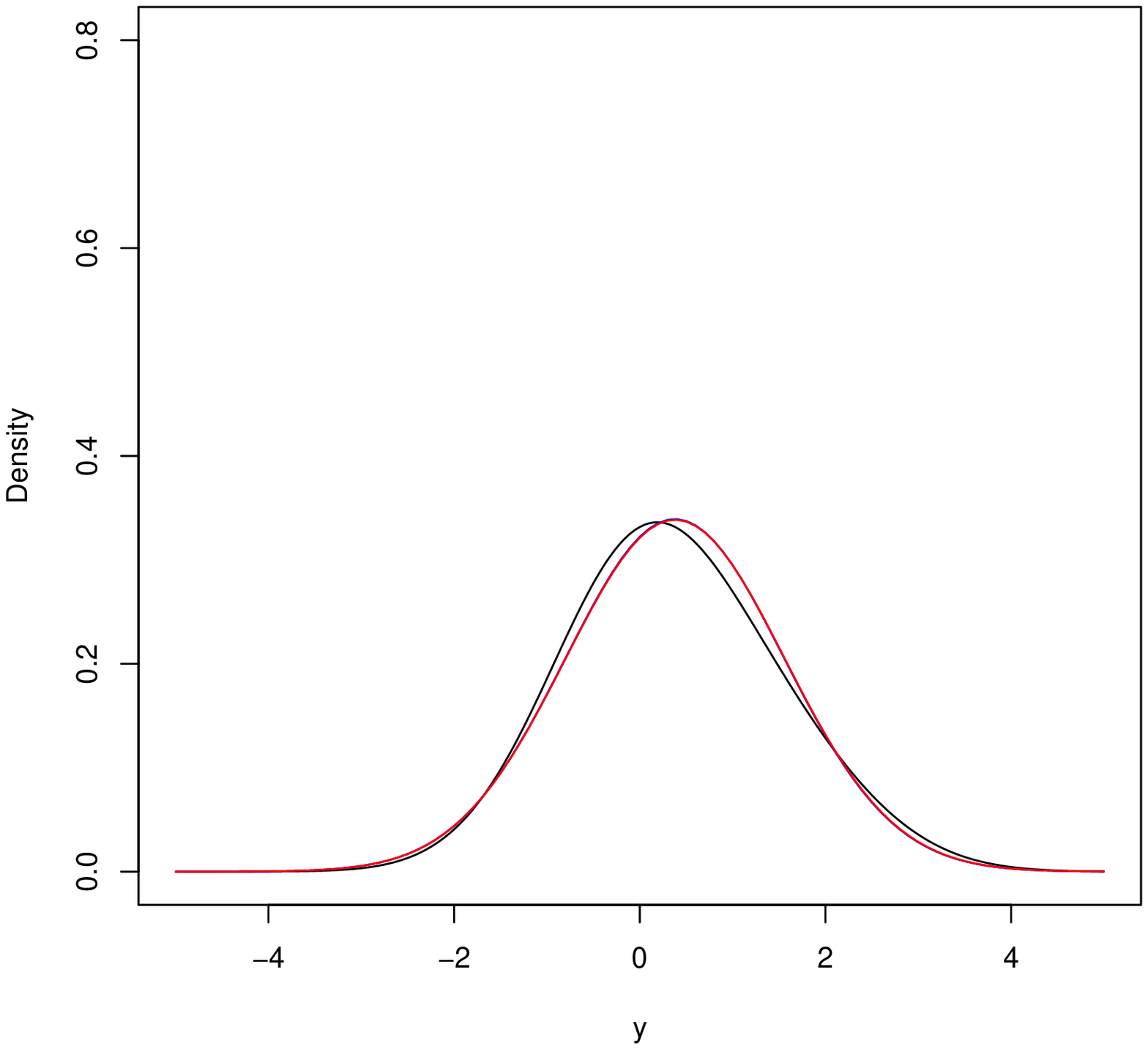}}
\subfigure[$\mu=2$]{\label{fig:mix1}\includegraphics[width=0.3\textwidth]{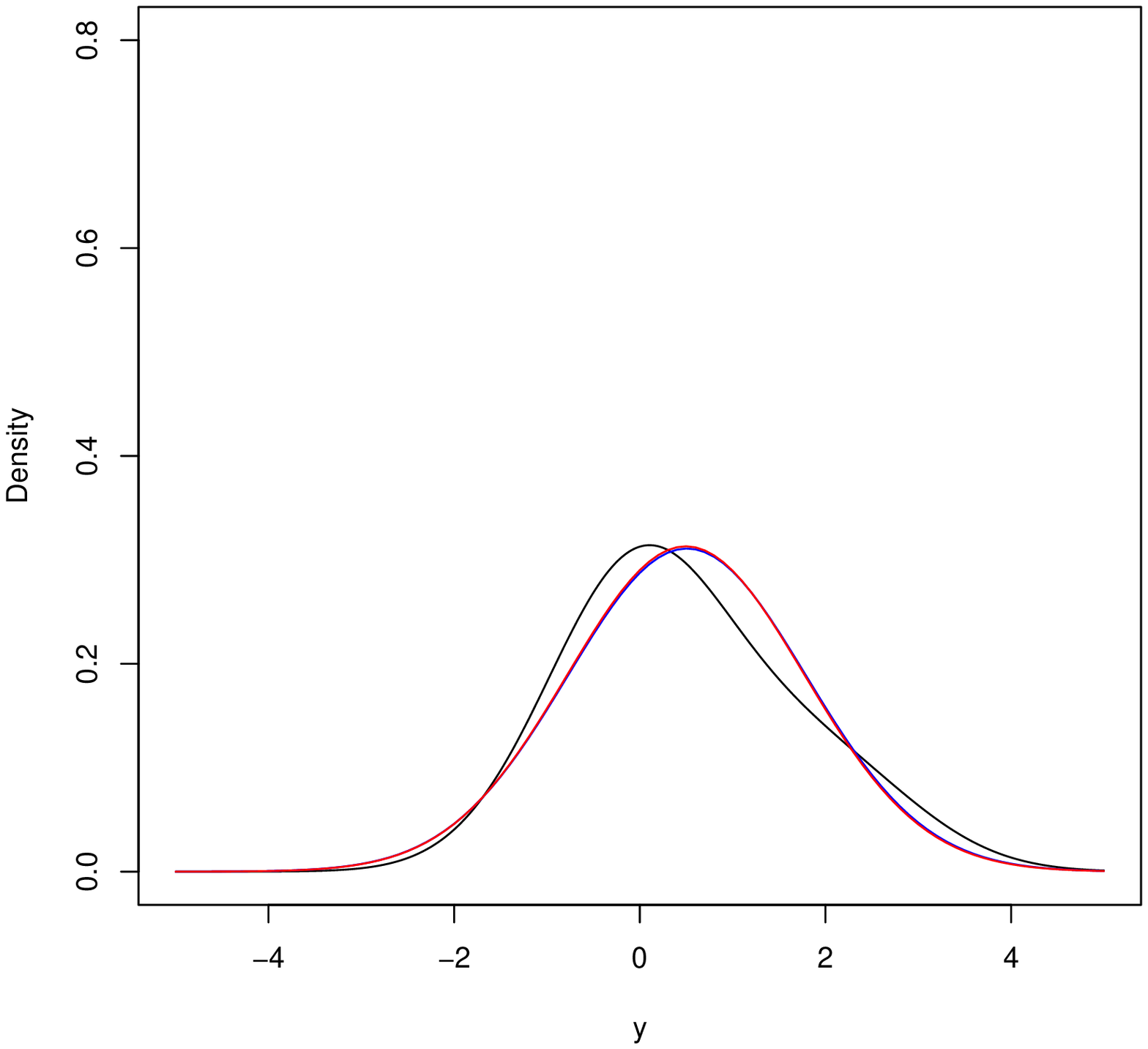}}
\subfigure[$\mu=2.5$]{\label{fig:mix1}\includegraphics[width=0.3\textwidth]{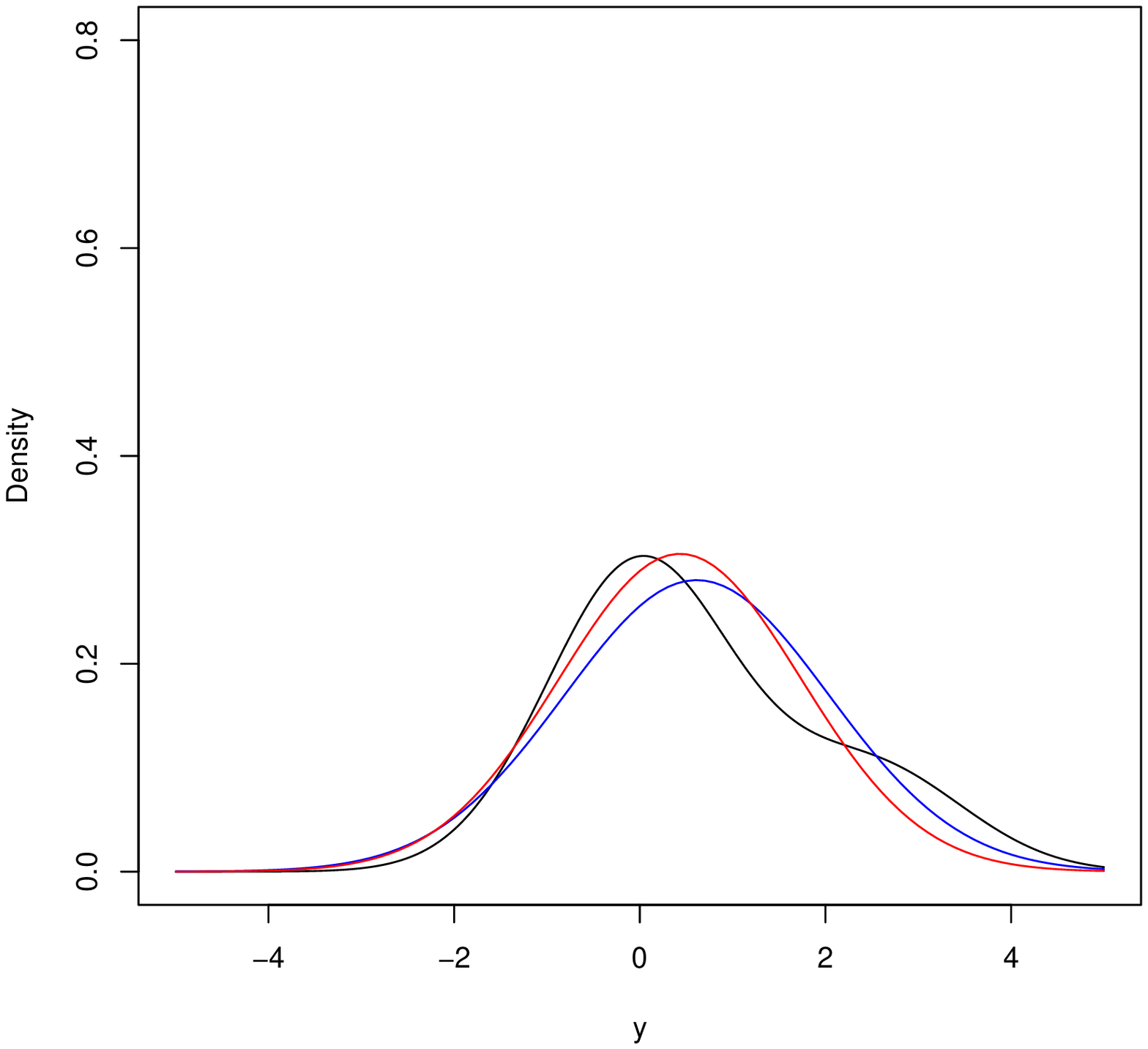}}
\end{center}
\caption{Approximation of a normal mixture density (black) $\frac{3}{4}\nm(0,1)+\frac{1}{4}\nm(\mu,1)$ by a normal based on minimizing Kullback--Leibler (blue) and Fisher divergence (red).}
\label{fig:mix}
\end{figure}
\begin{figure}[t]
\begin{center}
\subfigure[$\alpha=2$]{\label{fig:skew1}\includegraphics[width=0.3\textwidth]{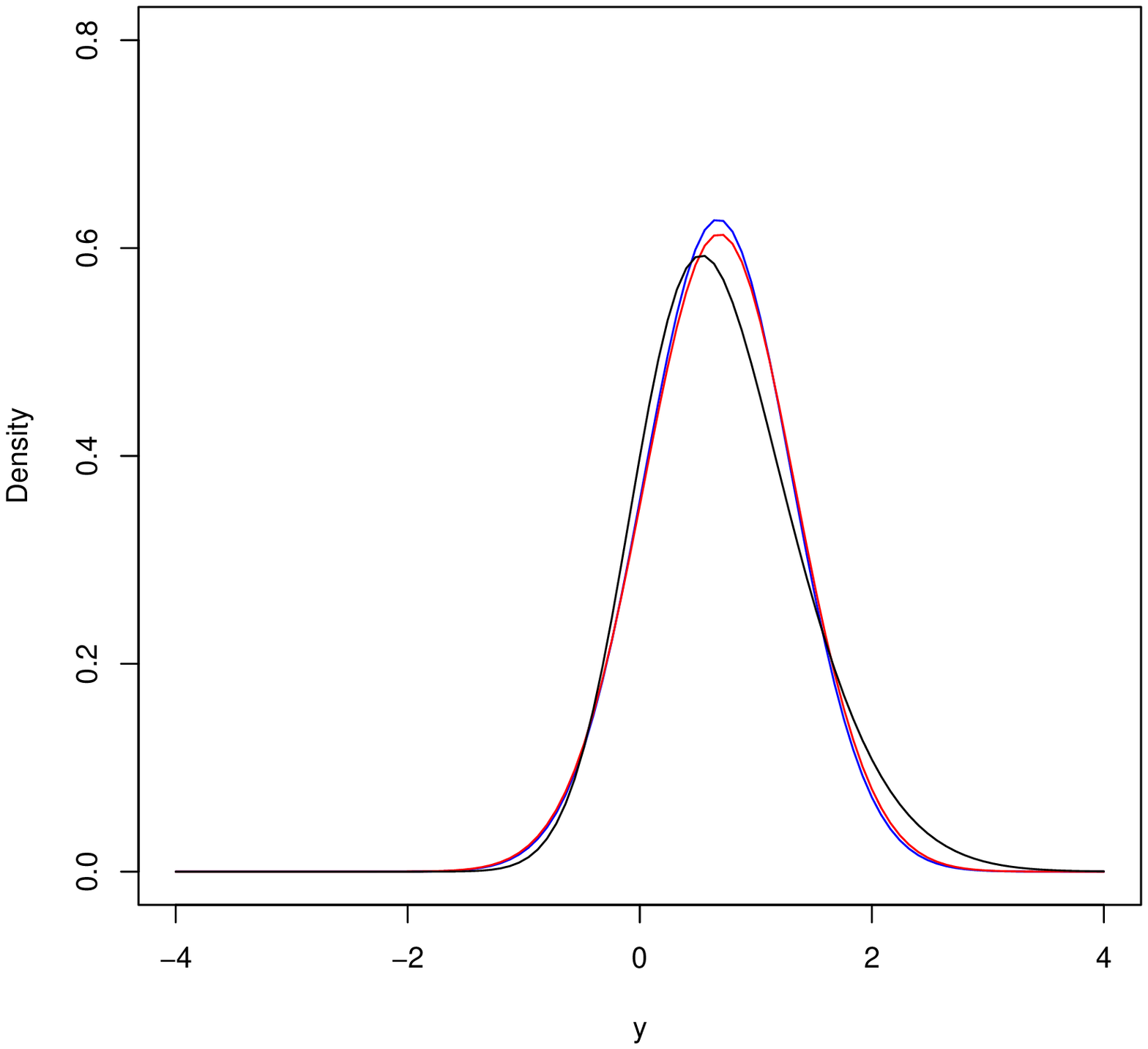}}
\subfigure[$\alpha=4$]{\label{fig:skew1}\includegraphics[width=0.3\textwidth]{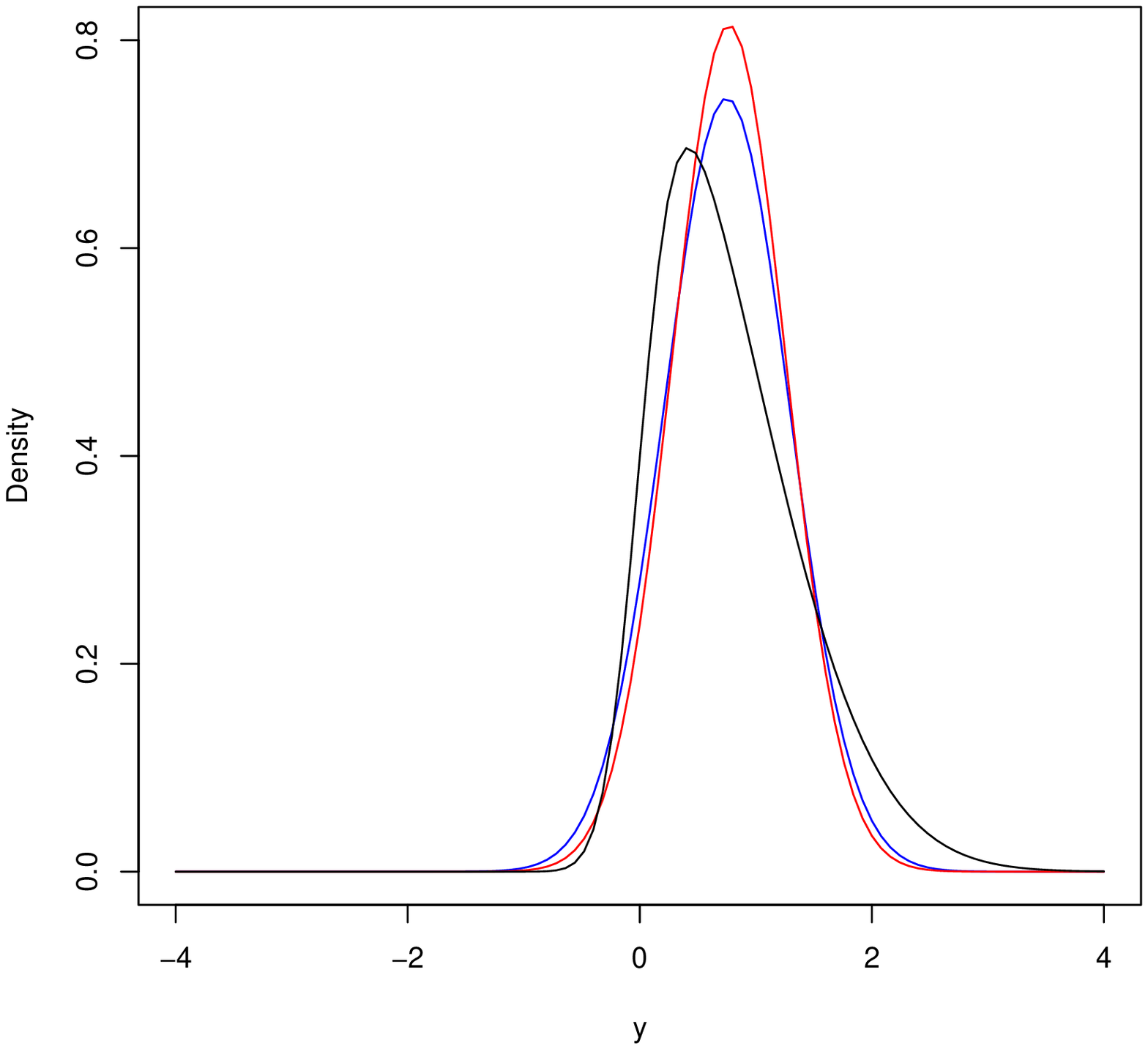}}
\subfigure[$\alpha=6$]{\label{fig:skew1}\includegraphics[width=0.3\textwidth]{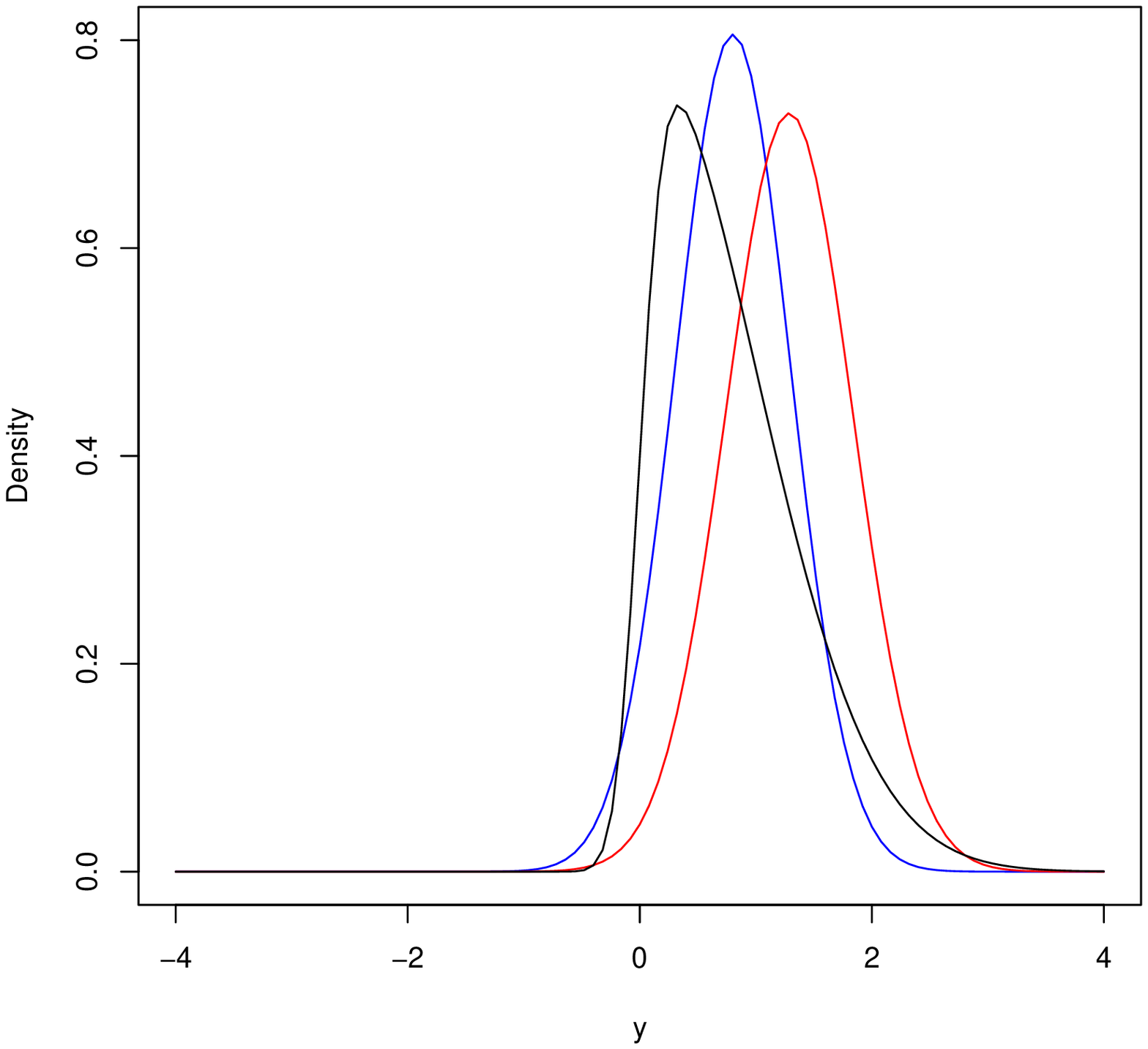}}
\end{center}
\caption{Approximation of a standard skew normal density (black) with shape $\alpha$ by a normal based on minimizing Kullback--Leibler (blue) and Fisher divergence (red).}
\label{fig:skew}
\end{figure}

Here consider three simple examples to illustrate the differences between approximations based on minimizing Kullback--Leibler and Fisher divergences.  The three examples where the true distribution is a Student-t, normal mixture, and skew normal consider the effect of tail heaviness, mild asymmetry, and skewness, respectively.  In all three cases, we consider approximations in the $\nm(\mu, \sigma^2)$ family.  Figures~\ref{fig:t}--\ref{fig:skew} show the results of solving these optimization problems.  For the heavy-tailed case in Figure~\ref{fig:t}, there is little difference between the two approximations for t-distribution with large degrees of freedom. For small $\nu$, the Fisher approximation has smaller variance than that based on Kullback--Leibler.  As for the mixture normal case in Figure~\ref{fig:mix}, two approximations perform almost the same for the first two mild asymmetry cases. In the third case, that Fisher approximation tends to capture the mode of the true distribution while the Kullback--Leibler approximation tries to capture the mean.  Finally, in the skew normal case, we find that when the skewness is mild, the approximations are mostly indistinguishable, but for fairly extreme skewness, there is a departure and, arguably, the Kullback--Leibler-based approximation is better in terms of moments; see Remark~\ref{re:huggins}.  This is because the Fisher divergence involves derivatives so the approximation will try to avoid regions where the true distribution is steeper than what the approximation family can accommodate.  While the Fisher-based approximation may not be universally better, but we find substantial improvements in the practically important logistic regression example in Section~\ref{S:results}.


\section{Variational approximation using $F$}
\label{S:approx}

Recall that the goal is to find the density $q^\star$ that minimizes the Fisher divergence, $F(q,\pi_y)$, of the posterior $\pi_y$ from the given family $\Qset$ of densities.  And recall that $F(q,\pi_y)$ does not depend on the potentially troublesome normalizing constant on the posterior; in fact, we could equivalently write $F(q,\tilde\pi_y)$.  

Consider a fairly general class $\Qset$ of approximating densities, namely, an exponential family of the form 
\begin{equation}
\label{eq:expfam}
q_\psi(\theta) = \exp\{\psi^\top s(\theta) + g(\psi) + h(\theta)\}, \quad \theta \in \Theta, 
\end{equation}
where $\psi \in \Psi \subseteq \RR^m$ is a free parameter, and $s(\theta)$, $g(\psi)$, and $h(\theta)$ are given functions.  Our proposal then is to take $\Qset = \{q_\psi: \psi \in \Psi\}$ as our approximating family.  Focusing on exponential family-based approximations is quite common in the literature \citep{blei.etal.vb.2017}.  Note, however, that we do not necessarily assume $\Qset$ to be a mean-field family, where the components of $\theta$ are modeled as independent.  We find this independence assumption to be quite limiting; indeed, the ability to more-or-less automatically capture relationships between components of $\theta$ is a major selling point of the Bayesian approach in general, and the mean-field approximation throws this away.  So working with an exponential family capable of modeling dependence is an advantage.  

Plugging density \eqref{eq:expfam} into the Fisher divergence formula yields
\begin{equation}
\label{eq:Fpsi}
F(\psi) := F(q_\psi, \pi_y) = \int \|z_y(\theta) - D(\theta) \psi\|^2 \, q_\psi(\theta) \, d\theta,
\end{equation}
where $z_y(\theta) = \grad \log \tilde\pi_y(\theta) - \grad h(\theta)$ and $D(\theta)$ is the $d \times m$ derivative matrix of the function $s$ in \eqref{eq:expfam} that maps $\RR^d$ to $\RR^m$.  The goal now is to find $\psi^\star = \arg\min F(\psi)$.

Notice that the only dependence on $\psi$ is in the linear term $D(\theta) \psi$ and in the density $p_\psi(\theta)$.  Consequently, $F(\psi)$ is almost quadratic in $\psi$, which suggests the following {\em iteratively re-weighted least squares} algorithm for computing the minimizer $\psi^\star$.  
\begin{enumerate}
\item[0.] Initialize $\psi^{(0)}$ and specify a convergence criterion; set $t=0$.
\item Fix $q_t=q_{\psi^{(t)}}$ and define 
\begin{equation}
\label{eq:ls}
\psi^{(t+1)} = \arg\min_{u \in \Psi} \int \|z_y(\theta) - D(\theta) \, u\|^2 \, q_t(\theta) \,d\theta. 
\end{equation}
\item If convergence criterion is satisfied, then return $\psi^\star = \psi^{(t+1)}$; otherwise, set $t \gets t+1$ and go back to Step~1. 
\end{enumerate}

In the unconstrained case, where $\Psi = \RR^m$, Step~1 can be written more explicitly.  That is, if we define  
\[ v_t = \int D(\theta)^\top z_y(\theta) \, q_t(\theta) \,d\theta \quad \text{and} \quad M_t = \int D(\theta)^\top D(\theta) \, q_t(\theta) \,d\theta, \]
then $\psi^{(t+1)} = M_t^{-1} v_t$ solves the quadratic problem. , Evaluating the function that maps $\psi$ to $(v, M)$ may not be too difficult.  In the relatively simple examples presented in Section~\ref{S:fisher}, this can be done with numerical integration.  Quadrature might not be feasible in higher-dimensional problems but, since the form of $q_\psi$ is known, Monte Carlo can be used to evaluate these expectations in certain cases.  Other strategies to approximate these integrals, e.g., Taylor approximation of the integrands, can also be employed.  

As a quick proof-of-concept, consider the case where $q_\psi(\theta) = \nm(\theta \mid \psi, \sigma^2)$ is a one-dimensional normal mean family with fixed variance $\sigma^2 > 0$.  Then the above algorithm's updates are as follows:
\[ \psi^{(t+1)} = \psi^{(t)} + \sigma^2 \int \grad \log \tilde\pi_y(\theta) \, \nm(\theta \mid \psi^{(t)}, \sigma^2) \, d\theta. \]
If there is a limit $\psi^\star$ as $t \to \infty$, then that limit must satisfy 
\[ \int \grad \log \tilde\pi_y(\theta) \, \nm(\theta \mid \psi^\star, \sigma^2) \, d\theta = 0. \]
Of course, if $\pi_y(\theta)$ is symmetric around its mean $\hat\theta_y$, then it follows that $\psi^\star = \hat\theta_y$.  Therefore, the proposed  approach aims to match the mean of the normal approximation with that of the symmetric posterior.  The far less trivial example in Section~\ref{S:results} show that the quality approximation properties in this normal case hold even more generally. 

However, iteratively re-weighted least squares algorithms are not guaranteed to converge.  To deal with such cases, we recommend using a weighted average of $M_t^{-1} v_t$ and $\psi^{(t)}$ to update $\psi$, i.e.,
\[ \psi^{(t+1)}=\rho \, M_t^{-1} v_t + (1-\rho)\, \psi^{(t)}, \quad \rho \in (0,1). \]
This modification of the iterative re-weighted least squares algorithm is used in \citet{karlovitz1970}, \citet{burrus1994iterative}, \citet{stone1991simplex}, and elsewhere. But it is worth noting that our weighted average update has a different motivation. Historically, this idea is used for $\ell_p$ approximation, when $p>2$, to achieve a more robust convergence in large $p$ cases. However, we only need to deal with $\ell_2$ approximation in the case of Fisher divergence, so our motivation is more closely aligned with that of momentum in gradient descent \citep[e.g.,][]{qian1999momentum}, which is to avoid oscillation and speed up convergence.

\section{Logistic regression application}
\label{S:results}

Consider a logistic regression model where $y=(y_1,\ldots,y_n)$ consists of independent observations, where $y_i \sim \ber(\eta_i)$ and 
\[ \text{logit}(\eta_i) = x_i^\top \theta, \quad i=1,\ldots,n. \]
Here $x_i$ is a known $d$-vector and $\theta \in \RR^d$ is an unknown vector of regression coefficients.  To complete the Bayesian model specification, take an independent prior, $\theta \sim \nm_d(0, \tau^2 I)$, where $\tau^2$ is set at 5 in our experiment.  Given the data and this model specification, it is possible to approximation the posterior distribution of $\theta$ using Markov chain Monte Carlo and here we employ a Metropolis--Hastings algorithm with $10^5$ iterations.  This approximation will be treated as the ``true'' posterior distribution to which our variational approximations will be compared.  

Since this is a regular exponential family model, we can expect that the posterior will look approximately normal. So it makes sense to consider a family of $d$-dimensional normal distributions 
\[ q_\psi(\theta) = \nm_d(\theta \mid \mu, \Sigma), \]
for the approximation, where $\mu \in \RR^d$ a free mean parameter and $\Sigma$ a free $d \times d$, symmetric, positive definite covariance matrix. Our approach, therefore, is to optimize over the $m$-vector $\psi$ consisting of the vector $\mu$ of $d$ means and the $d(d+1)/2$ entries on and above the diagonal of $\Sigma$, so that $m=d(d+3)/2$.  

Despite its apparent simplicity, the fact that this model is non-conjugate makes applying the traditional variational methods based on Kullback--Leibler divergence a challenge.  Indeed, various extensions to the traditional framework have been proposed to deal with non-conjugacy  \citep[e.g.,][]{braun2010variational, wang2013variational}, along with several ``black box'' methods \citep[e.g.,][]{AutoVB, ranganath2014black}.  But our proposed Fisher divergence-based approximation can be applied direct to conjugate and non-conjugate models alike.  For our simulation experiments here, in addition to our proposed approximation, we consider the method in \citet[JJ,][]{jaakkola1997} that is designed specifically for the logistic regression setting and the doubly stochastic variational inference method in \citet[DSVI,][]{titsias2014doubly}.

For our experiments, we set $p=5$ and vary $n$ from 100, 200, to 500.  We fix the prior variance at $\tau^2=5$.  Each data set is generated by first simulating the true $\theta$ from an isotropic Gaussian distribution with mean 0 and variance 1.  Then the $x_i$'s are sampled from either an isotropic Gaussian distribution with mean 0 and variance 3, or from a first-order autoregressive model with correlation parameter 0.8.  Given $\theta$ and the $x_i$'s, the $y_i$'s are generated according to the logistic regression model described above.  For each of the two kinds of covariate models, 100 data sets are generated.  

Tables~\ref{table:iso} and \ref{table:ar} summarize the estimation accuracy of each of the variational approximation methods relative to the ``truth'' based on Markov chain Monte Carlo for the two covariate distributions, respectively.  In particular, we report the averages of $\|\hat\mu - \mu_{\text{\sc mcmc}}\|$ and $\|\widehat \Sigma - \Sigma_{\text{\sc mcmc}}\|_F$, where $\|A\|_F = \{\text{tr}(A^\top A)\}^{1/2}$ is the Frobenius norm.  While all three variational approaches have similar performance in estimating the posterior mean, our Fisher divergence-based method is the most accurate in terms of covariance matrix estimation in all of the cases; see Remark~\ref{re:huggins}.  To visualize the methods' performance, Figure~\ref{fig:marginal} shows marginal posterior distributions for $(\theta_2,\theta_4)$ and $(\theta_1,\theta_3)$ for a single data set of size $n=200$ under the isotropic covariate setting.  Here we can see that while the contour plots based on our Fisher approximation are quite similar to the true posterior's contours, those based on JJ are much narrower and DSVI incorrectly estimates some of the correlations in this dataset.  

  \begin{table}[t]
	\centering
	\begin{tabular}{c c c c }
		\hline
		$n$&Method& $\|\hat\mu - \mu_{\text{\sc mcmc}}\|$ & $\|\widehat \Sigma - \Sigma_{\text{\sc mcmc}}\|_F$ \\
		\hline
  	100&Fisher&0.885 &0.338\\
		&JJ&1.020&0.672\\
		&DSVI&0.826 &0.392 \\
	   \hline
	   200&Fisher& 0.150&0.024 \\
	   &JJ& 0.233& 0.151\\
	   &DSVI& 0.106& 0.150\\
	 \hline
      500&Fisher&0.039 &0.003\\
      &JJ& 0.073&0.041\\
      &DSVI& 0.024& 0.076\\
      \hline
  \end{tabular}
\caption{Comparison of the variational approximations in terms of posterior mean and covariance matrix estimation error in the isotropic covariate case.}
  \label{table:iso}
  \end{table}
  
   \begin{table}[t]
	\centering
	\begin{tabular}{c c c c}
		\hline
		$n$&Method& $\|\hat\mu - \mu_{\text{\sc mcmc}}\|$ & $\|\widehat \Sigma - \Sigma_{\text{\sc mcmc}}\|_F$ \\
		\hline
  	100&Fisher&0.804 &0.371\\
		&JJ&0.922&0.736\\
		&DSVI&0.759 &0.504 \\
	  \hline
	  200& Fisher& 0.133&0.027 \\
	   &JJ& 0.205& 0.157\\
	   &DSVI& 0.114& 0.177\\
      \hline
      500&Fisher&0.051 &0.006\\
      &JJ& 0.084&0.056\\
      &DSVI& 0.041& 0.111\\
      \hline
  \end{tabular}
\caption{Comparison of the variational approximations in terms of posterior mean and covariance matrix estimation error in the autoregressive covariate case.}
  \label{table:ar}
  \end{table}
\begin{figure}[t]
 \label{fig:marginal}
\begin{center}
\subfigure{\label{fig:marginal1}\includegraphics[width=0.23\textwidth]{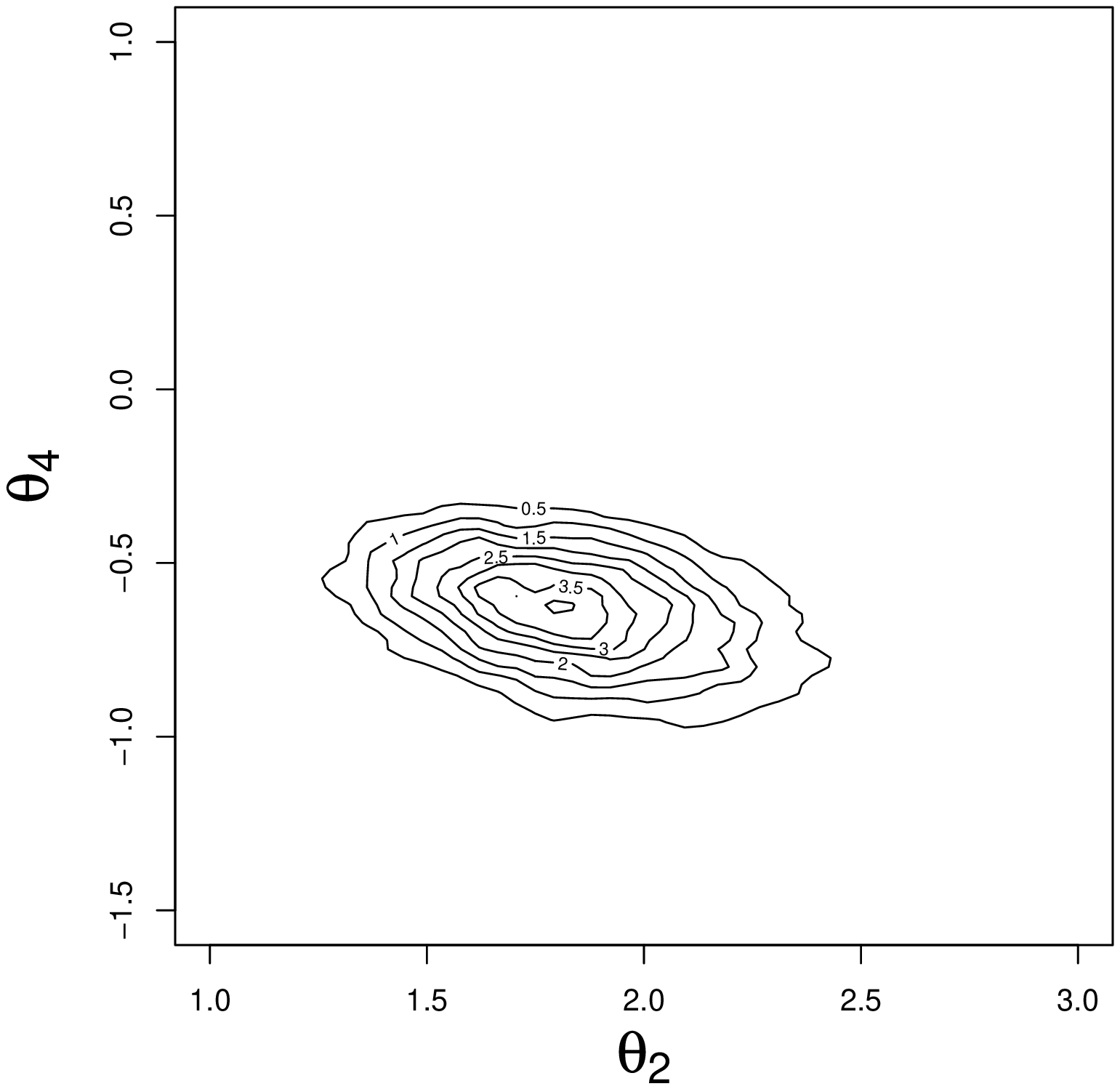}}
\subfigure{\label{fig:marginal2}\includegraphics[width=0.23\textwidth]{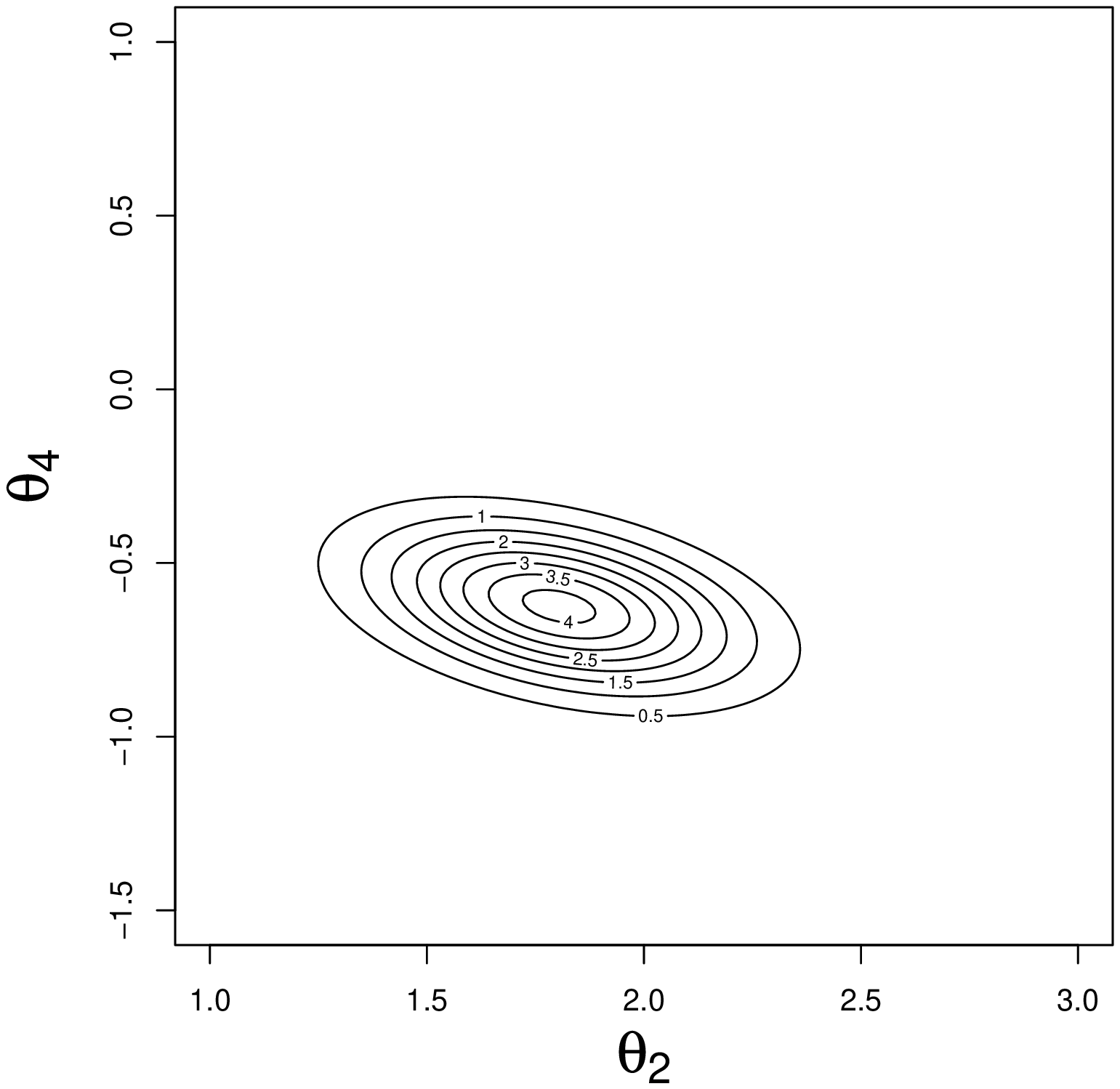}}
\subfigure{\label{fig:marginal3}\includegraphics[width=0.23\textwidth]{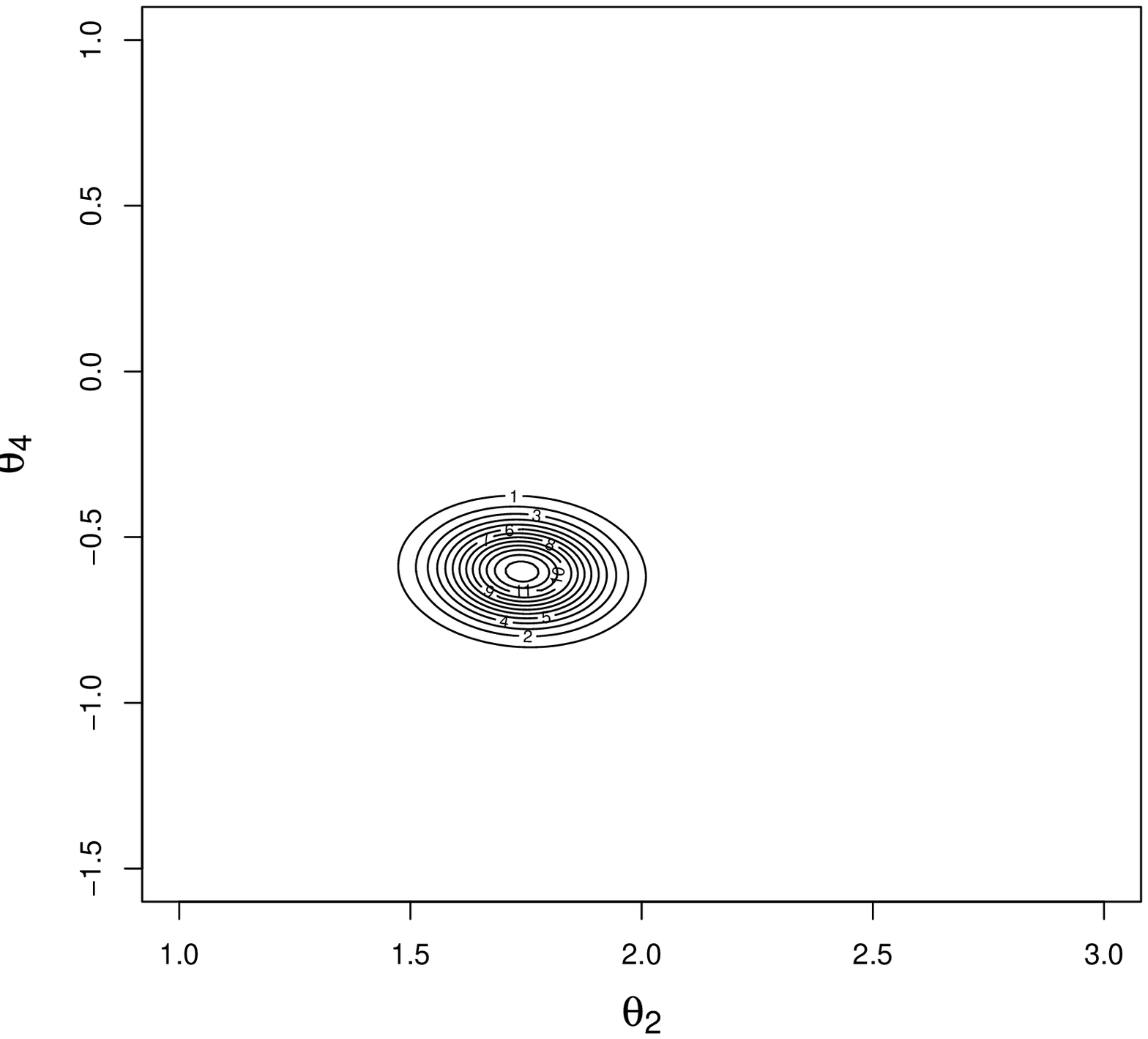}}
\subfigure{\label{fig:marginal4}\includegraphics[width=0.23\textwidth]{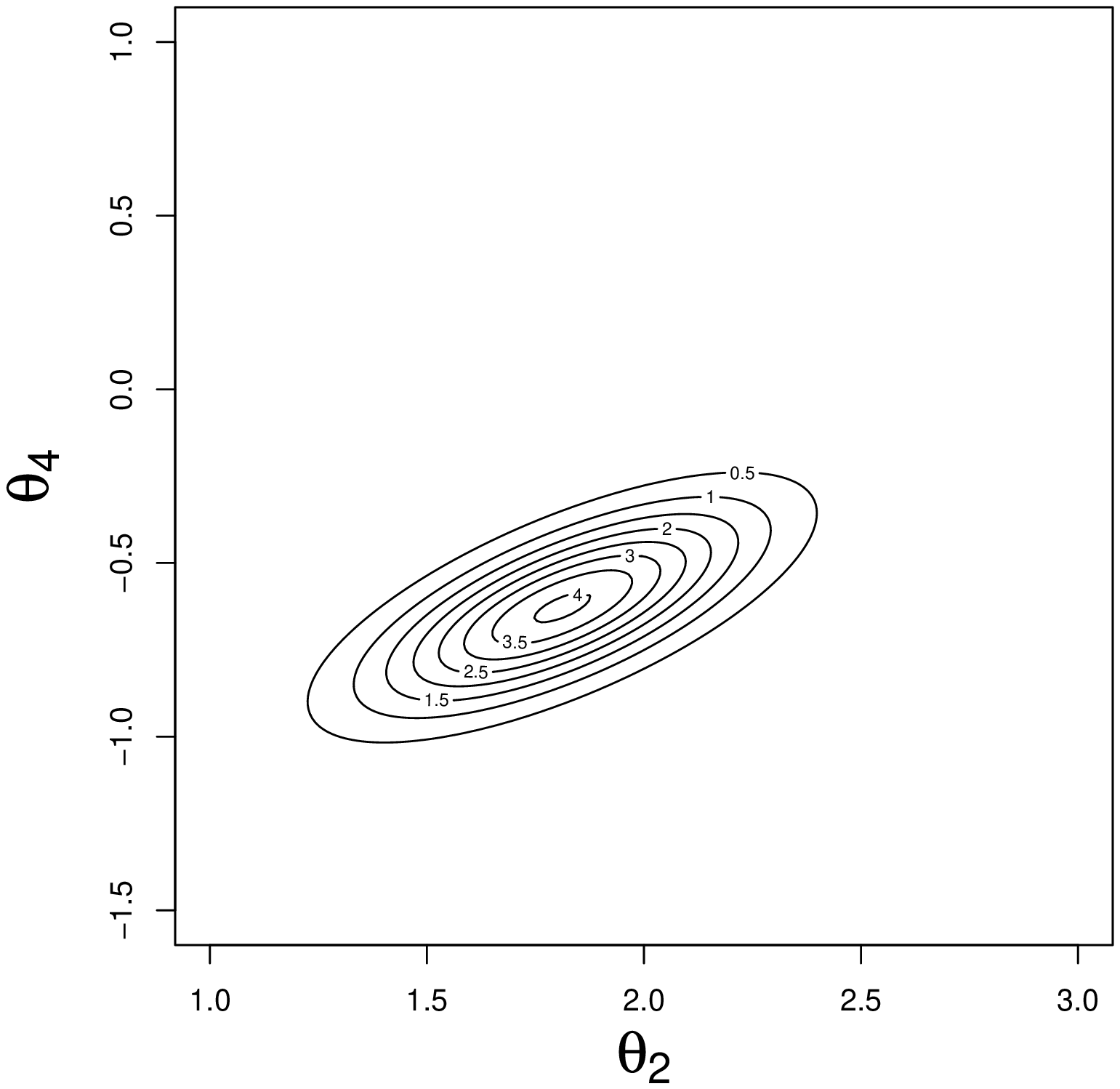}}

\subfigure{\label{fig:marginal5}\includegraphics[width=0.23\textwidth]{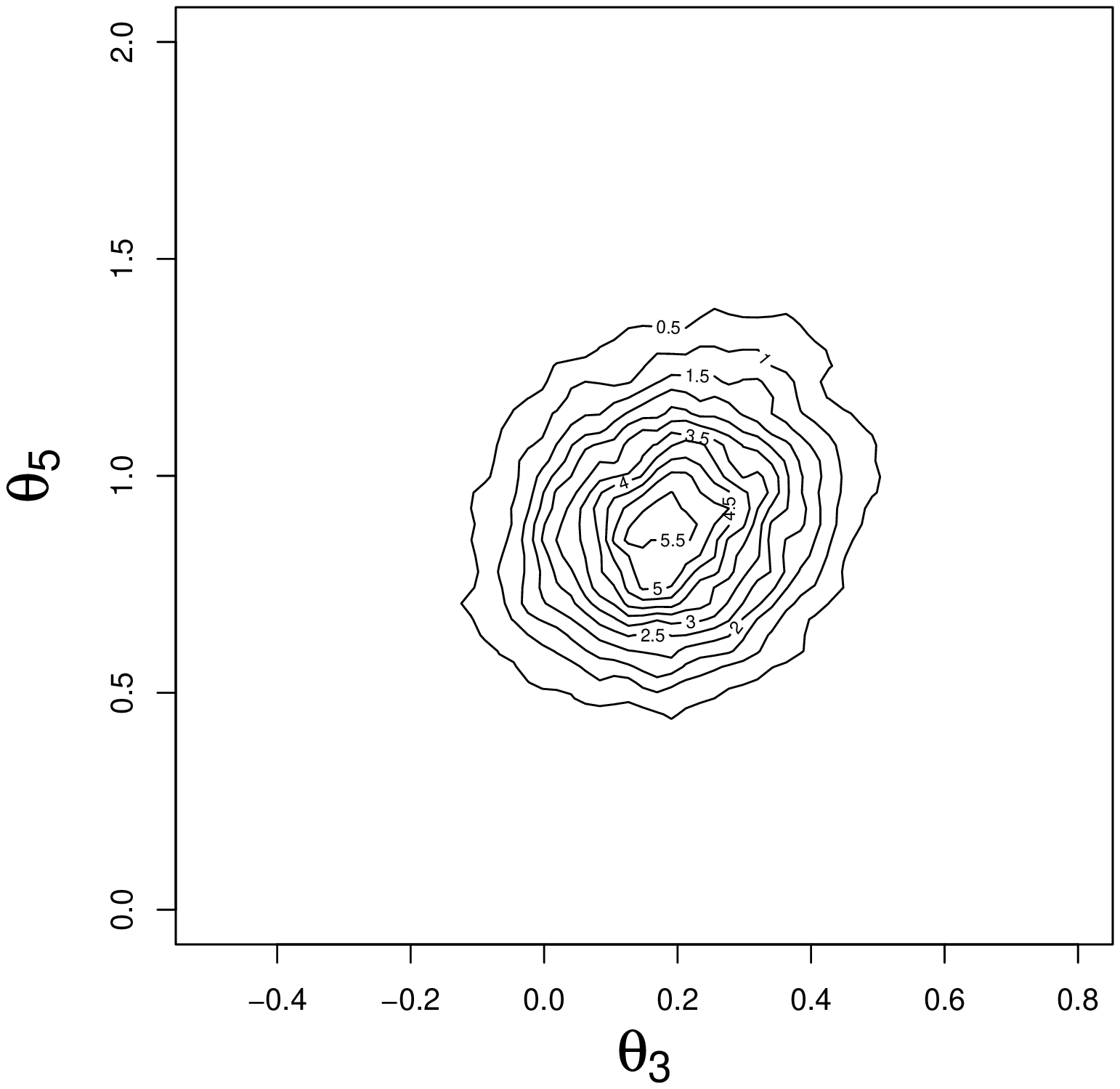}}
\subfigure{\label{fig:marginal6}\includegraphics[width=0.23\textwidth]{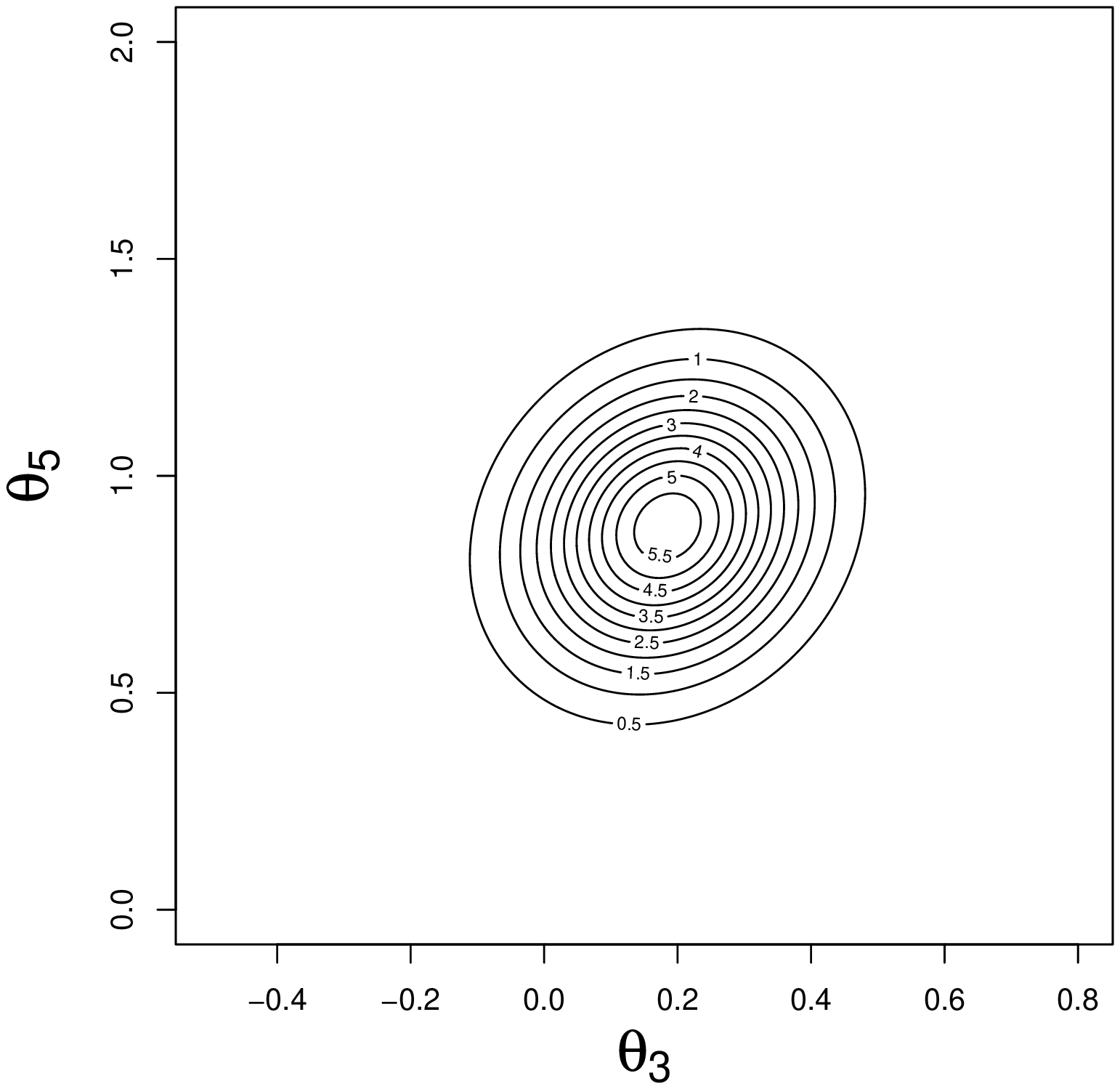}}
\subfigure{\label{fig:marginal7}\includegraphics[width=0.23\textwidth]{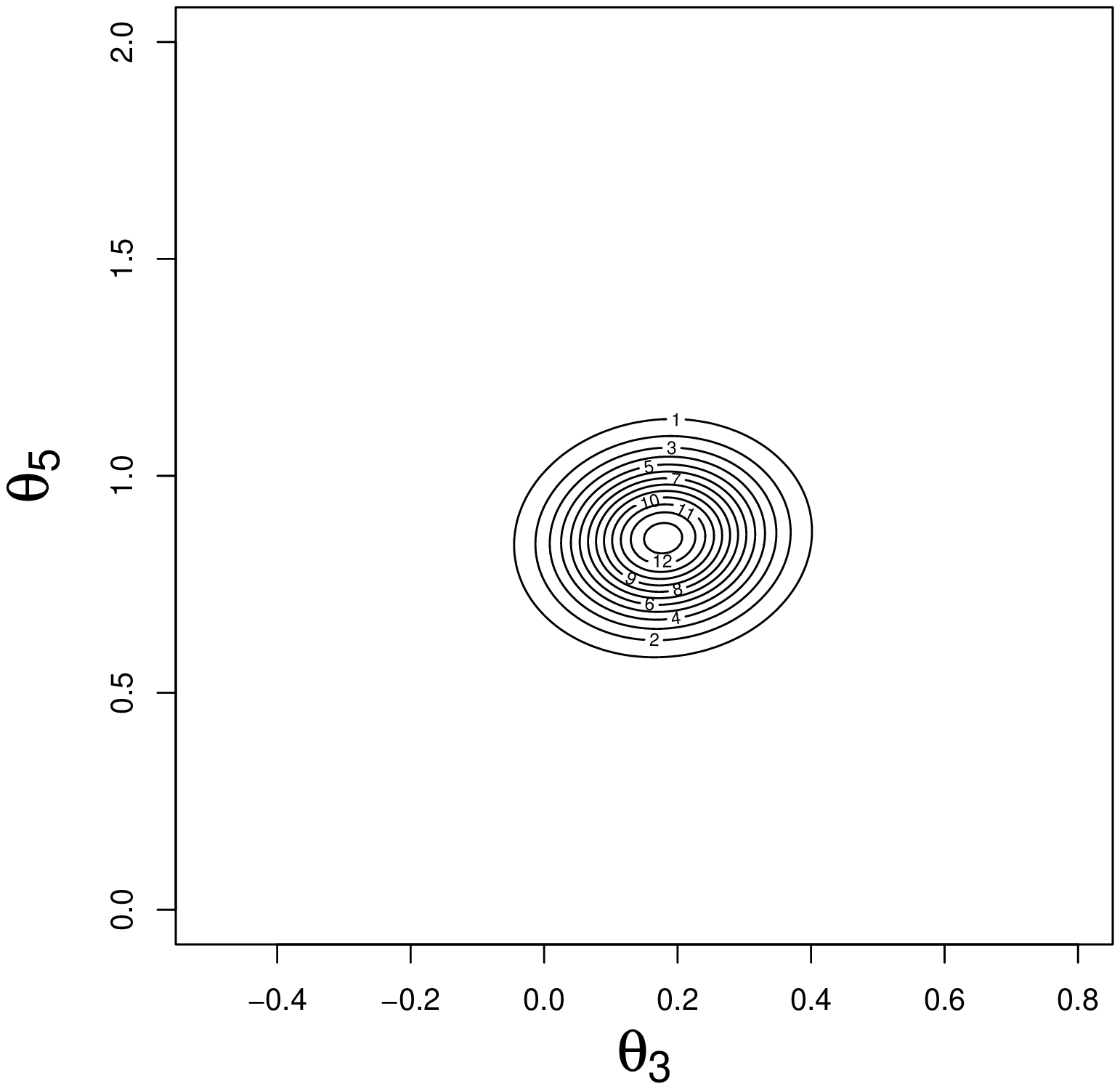}}
\subfigure{\label{fig:marginal8}\includegraphics[width=0.23\textwidth]{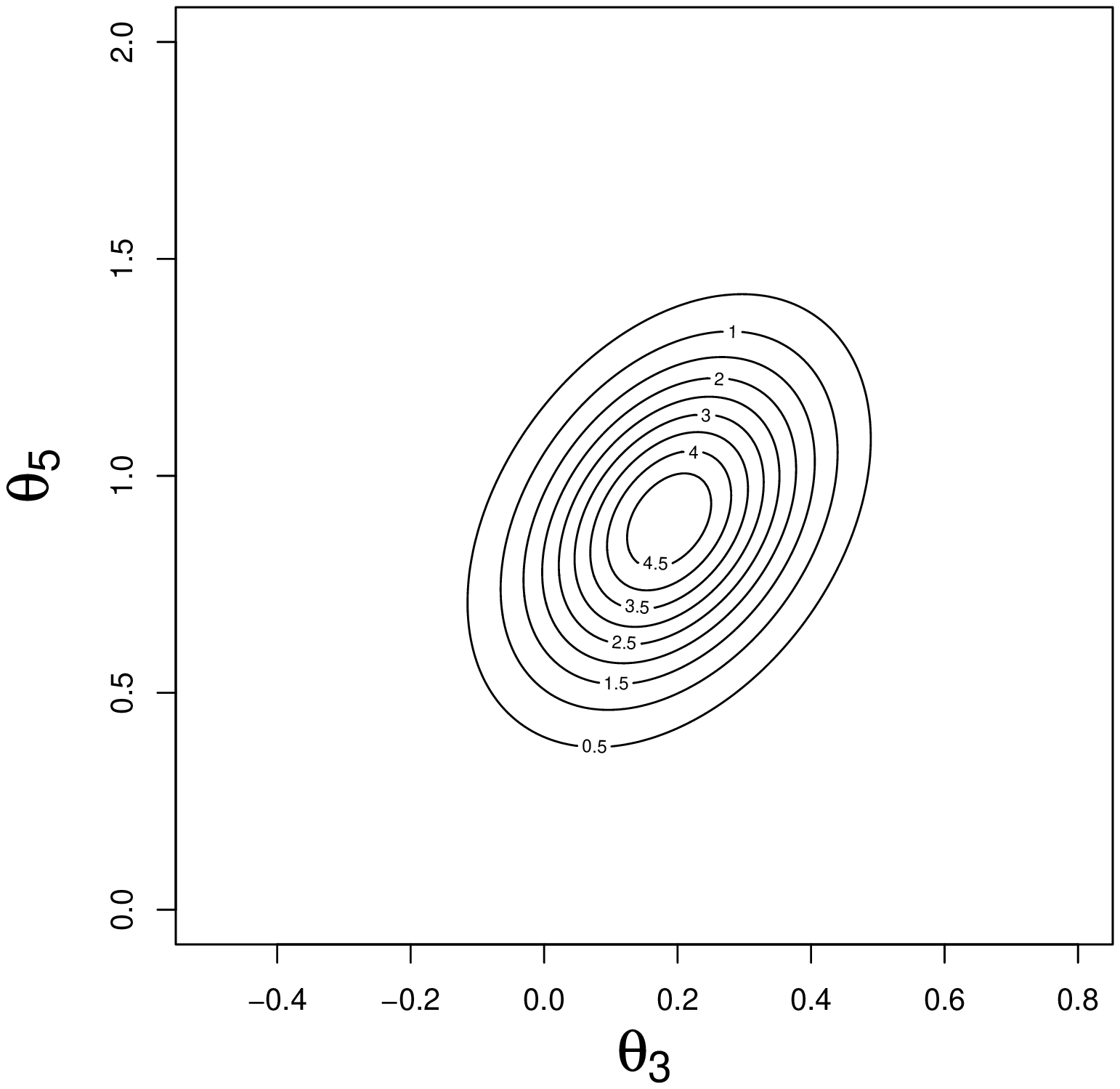}}
\end{center}
\caption{The first and second rows show contour plots of two different pairs' marginal distributions, namely, $(\theta_2,\theta_4)$ and $(\theta_3,\theta_5)$, respectively. The four columns correspond to MCMC, Fisher, JJ, and DSVI, respectively.}
\label{fig:marginal}
\end{figure}

\begin{figure}[t]
\begin{center}
\subfigure[$n=100$, isotropic]{\label{fig:coverage1}\includegraphics[width=0.29\textwidth]{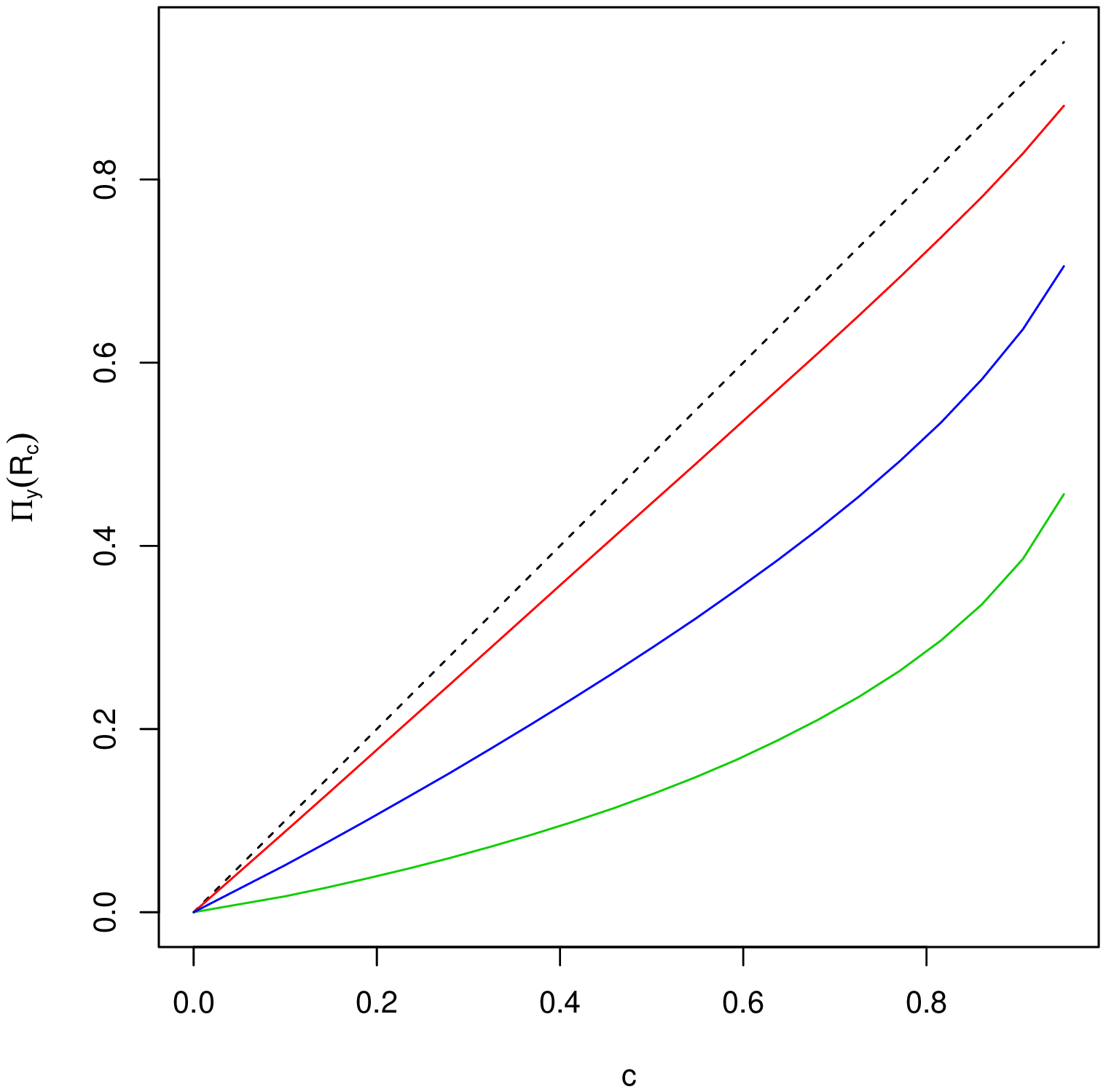}}
\subfigure[$n=200$, isotropic]{\label{fig:coverage2}\includegraphics[width=0.29\textwidth]{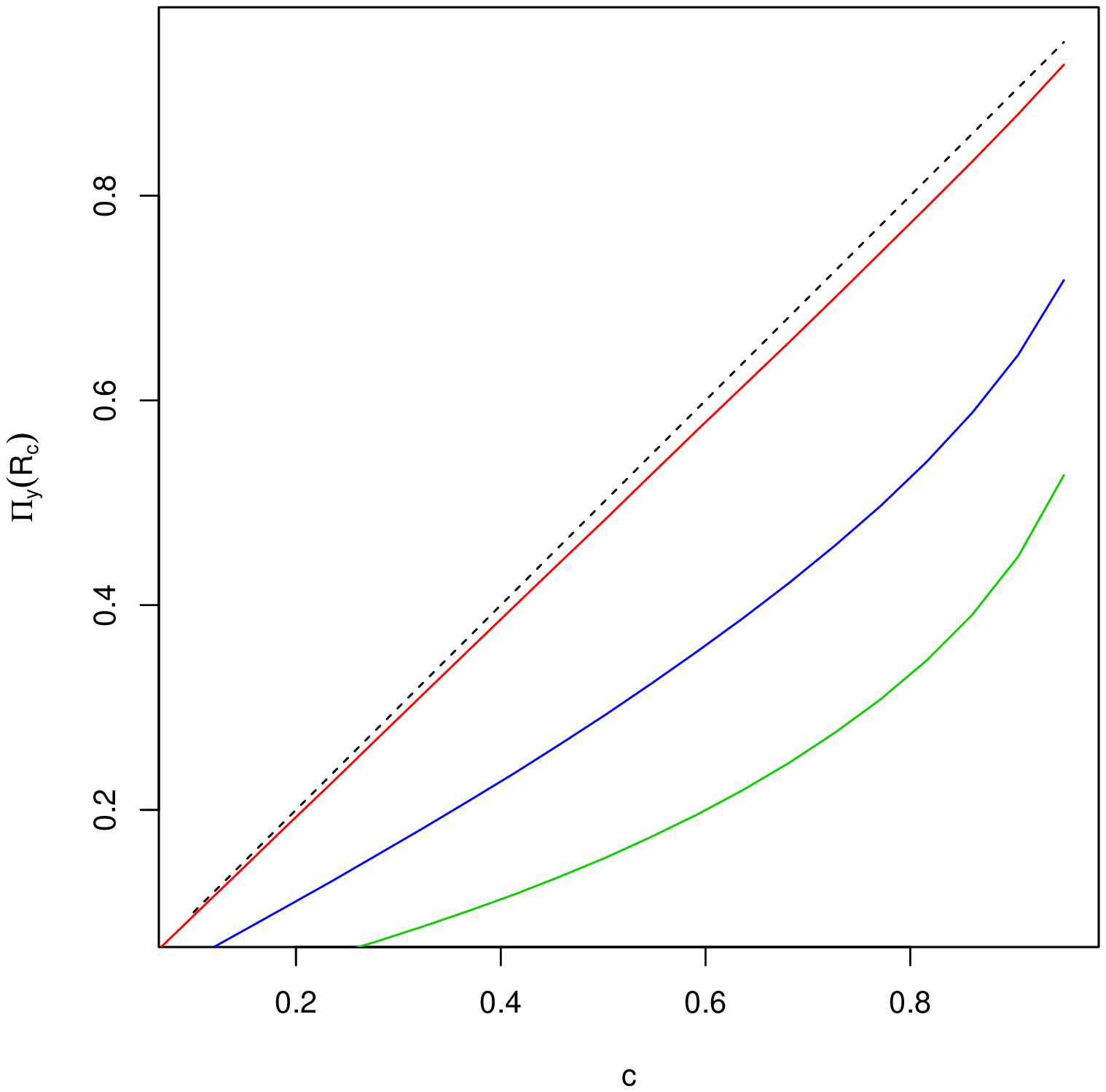}}
\subfigure[$n=500$, isotropic]{\label{fig:coverage3}\includegraphics[width=0.29\textwidth]{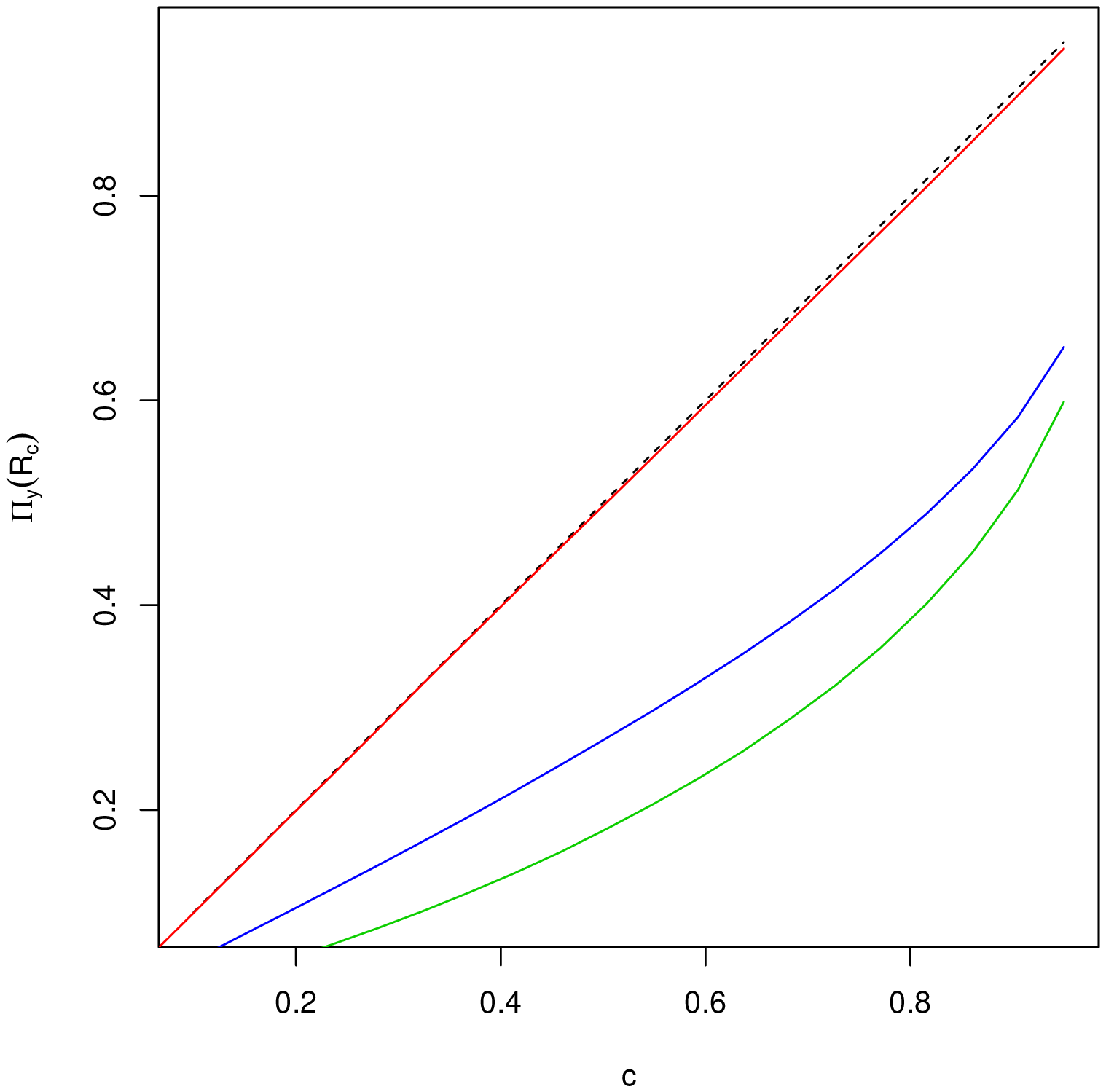}}

\subfigure[$n=100$, AR(1)]{\label{fig:coverage4}\includegraphics[width=0.29\textwidth]{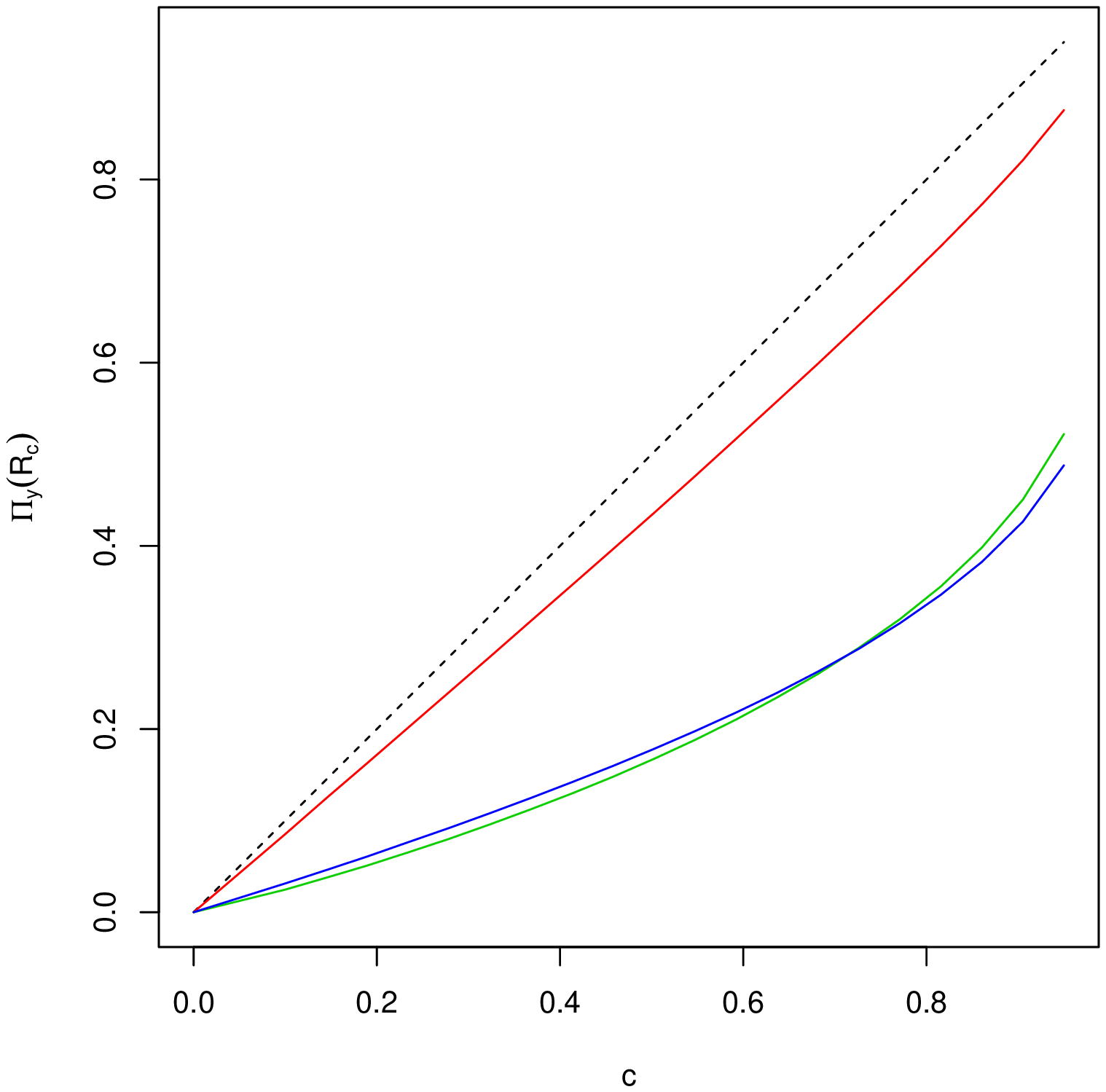}}
\subfigure[$n=200$, AR(1)]{\label{fig:coverage5}\includegraphics[width=0.29\textwidth]{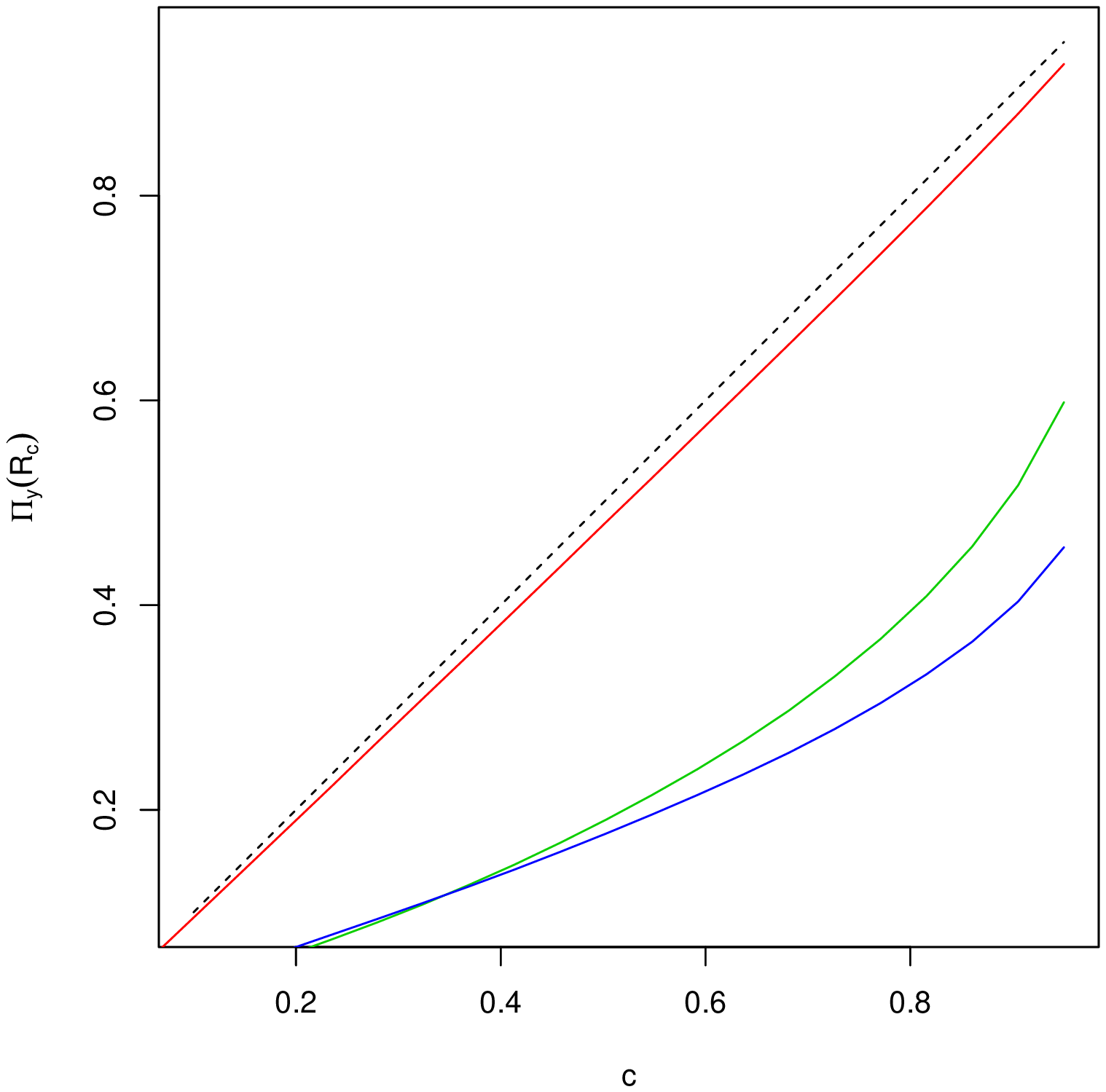}}
\subfigure[$n=500$, AR(1)]{\label{fig:coverage6}\includegraphics[width=0.29\textwidth]{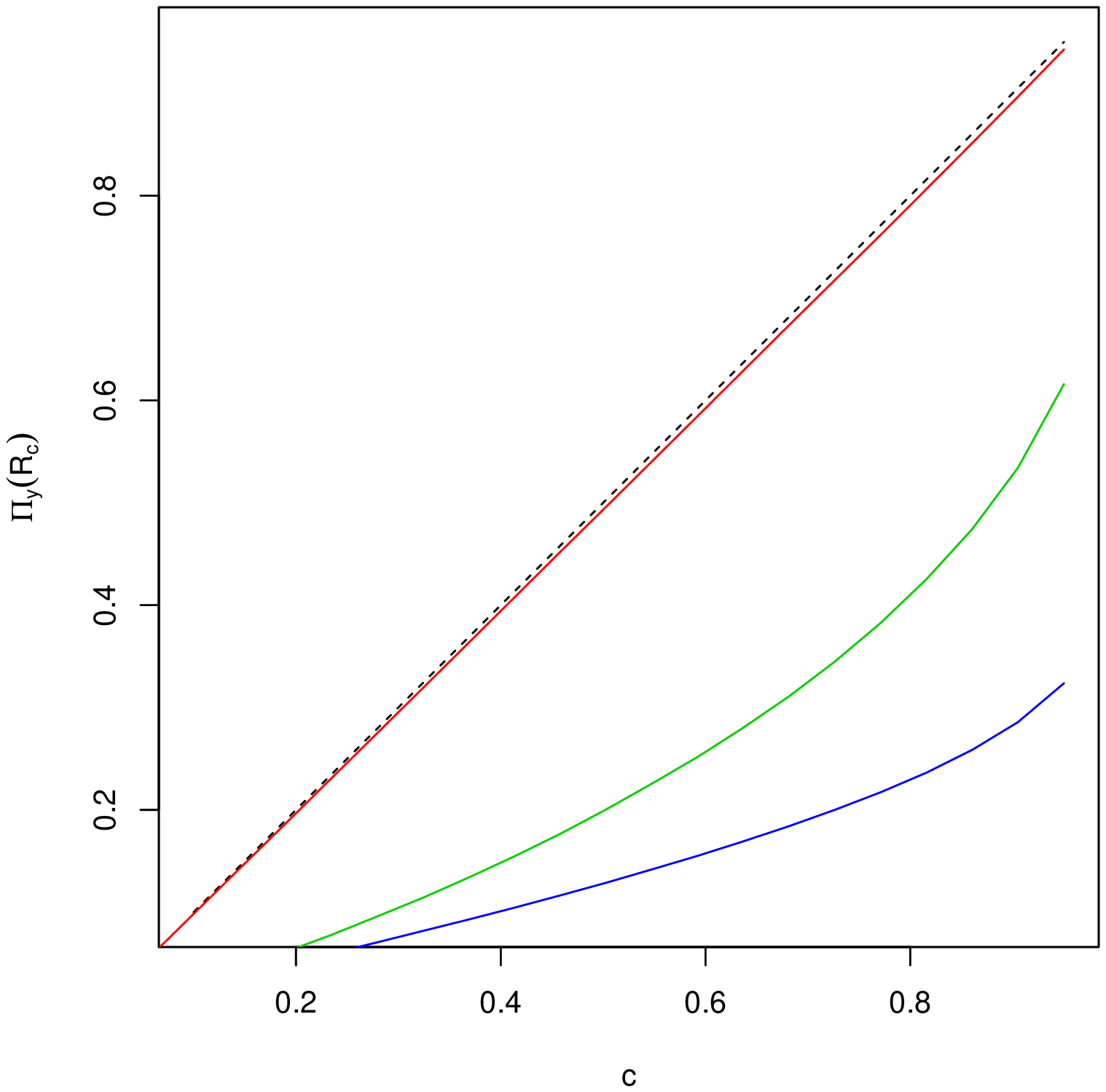}}
\end{center}
\caption{$R_c$ is the $100c$\% credible region based on the Fisher (red), JJ (green), and DSVI (blue) approximation, and $\Pi_y(R_c)$ is the corresponding posterior probability.}
\label{fig:coverage}
\end{figure}

To better understand how close the contours of the approximations match up to those of the true posterior, we compute the posterior probability, $\Pi_y(R_c)$, of the $100c$\% credible regions/contours, $R_c$, as $c$ varies in $(0,1)$.  If the approximation is accurate, then $R_c$ should have posterior probability close to $c$, so $c \mapsto \Pi_y(R_c)$ should be close to the 45-degree line; otherwise, the curve will be above or below the 45-degree line, depending on whether the approximation over- or under-estimates the posterior dispersion.  According to Figure~\ref{fig:coverage}, our proposed Fisher approximation has contours that reflect those of the true posterior distribution much more accurately than the other variational methods.  And our high-quality approximation of the true posterior is achieved in just a matter of seconds---compared to minutes for MCMC---which is only slightly slower than the very-fast but less-accurate JJ method.

\section{Discussion} 
\label{S:discuss}

In this paper, we proposed a new variational approximation based on the Fisher divergence. Compared with more traditional variational methods based on minimizing the Kullback--Leibler divergence, this new method does not make mean-field/independence assumptions about the components of $\theta$, and can be applied to any class of models, conjugate or not.  The iteratively re-weighted least squares algorithm in Section~\ref{S:approx} is straightforward and the method yields approximations that are very accurate compared to existing methods in the benchmark logistic regression application.

While our proposed Fisher divergence-based approximation is orders of magnitude faster than Markov chain Monte Carlo, there are still opportunities to improve the efficiency.  In most applications, including logistic regression, evaluating the integrals involved in solving the least squares problem in \eqref{eq:ls} may require Monte Carlo, which can slow down the computations.  One should be able to employ a version of stochastic gradient descent to solve the optimization problem more efficiently.  

There are also a couple interesting theoretical questions to be investigated.  First, in traditional variational approaches, the log-marginal likelihood appears as an additive constant in the Kullback--Leibler divergence, and manipulating this expression results in a bound---the so-called ``evidence lower bound''---which can be used for model comparison purposes, etc.  Here, when working with the Fisher divergence, the marginal likelihood disappears, so no bound is immediately available.  However, as noted in Section~\ref{S:fisher}, there are connections between the Fisher divergence and other discrepancy measures, so perhaps these connections can be exploited to derive a bound on the marginal likelihood.  Second, as we pointed out in Remark~\ref{re:huggins}, the results in \citet{huggins.etal.fisher} suggest that minimizing Fisher divergence will provide improved error control on the estimates of certain posterior moments compared to working with Kullback--Leibler divergence, but this is not necessary.  This is because practicality requires that we write the Fisher divergence as an expectation with respect to the {\em approximating distribution}, not the true posterior, and this difference in measure affects the quality of the bounds in \citet{huggins.etal.fisher}.  So it remains to understand in what situations their bounds will translate to improved moment estimates when using Fisher compared to Kullback--Leibler divergence.


\appendix

\section{Fisher updates in logistic regression}
For the logistic regression case Section~\ref{S:results}, define $\Omega=\Sigma^{-1}=(\omega_{ij})_{i,j=1,\dots,d}$ be the precision matrix of the normal variational approximation. The optimization vector is $$\psi=(-0.5\omega_{11},\dots,-0.5\omega_{dd},-\psi_1,\dots,-\psi_{d-1},(\Omega \mu)^\top)^\top,$$ 
where $\psi_i=(\omega_{1,1+i},\omega_{2,2+i},\dots,\omega_{d-i,d})^\top$, $i=1,\dots,d-1$.   The corresponding $v^{(t)}$ is
\begin{equation}
\label{equ:vt}
    \begin{aligned}
   v^{(t)}=\begin{pmatrix}
    a_0^{(t)}\\
    \vdots\\
    a_{d-1}^{(t)}\\
    \sum_{i=1}^n \bigl[ y_i-1+\E_{q_t}\bigl\{ \frac{1}{1+\exp(x_i^\top \theta)} \bigr\} x_{i1}-\sigma^{-2} \mu_1^{(t)} \bigr] \\
    \vdots\\
     \sum_{i=1}^n \bigl[ y_i-1+\E_{q_t}\bigl\{ \frac{1}{1+\exp(x_i^\top \theta)} \bigr\} x_{id}-\sigma^{-2} \mu_d^{(t)} \bigr]
    \end{pmatrix}
    \end{aligned},
\end{equation}
where $a_k^{(t)}$ is a $(d-k)$-vector, for $k=0,1,\ldots,d-1$, with $j^{\text{th}}$ entry equal to 
\[  \sum_{i=1}^n \Bigl[(y_i-1)(\mu_{j+k}^{(t)}x_{i,j}+\mu_j^{(t)}x_{i,j+k})+\E_{q_t}\Bigl\{\frac{x_{i,j}\theta_{j+k}+x_{i,j+k}\theta_j}{1+\exp(x_i^\top \theta)}\Bigr\} \Bigr]-\frac{2(\mu_j^{(t)}\mu_{j+k}^{(t)}+\omega_{j,j+k}^{(t)})}{\sigma^2}. \]
The matrix $M^{(t)}$ is symmetric and has a block sparse structure that consists of quadratic terms and crossproducts of $\theta$; the expressions are messy and not worth writing down here.  

In $v^{(t)}$, we need to compute $\E_{q_t}\{ f_{ij}(\theta) \}$, where 
\[ f_{ij}(\theta) = \frac{\theta_j}{1 + \exp(x_i^\top \theta)}, \quad i=1,\ldots,n, \quad j=1,\ldots,d. \]
Since there is no explicit form for this expectation and it is time-consuming to estimate it via Monte Carlo, we propose the following approximation.  Write out a second order Taylor series expansion around variational distribution mean $\mu$, i.e., 
\begin{equation}
\label{taylor}
  f_{ij}({\theta}) \approx f_{ij}(\mu)+\nabla f_{ij}(\mu) (\theta-\mu)+\tfrac{1}{2}(\theta-\mu)^\top H_{ij}(\mu)(\theta-\mu),
\end{equation}
where $\nabla f_{ij}(\mu)$ is the gradient (a row vector) and $H_{ij}(\mu)=\nabla^2f_{ij}(\theta)|_{\theta=\mu}$ is the Hessian matrix of $f_{ij}(\theta)$, both evaluated at $\theta=\mu$.  With the approximation in \eqref{taylor}, and since $\E_{q_t}(\theta) = \mu$, we approximate the expectation as 
\[ \E_{q_t}\{f_{ij}(\theta)\} \approx f_{ij}(\mu)+\tfrac{1}{2}\E_{q_t}\{(\theta-\mu)^\top H_{ij}(\mu)(\theta-\mu)\}. \]
Again, since $\theta \sim \nm(\mu,\Sigma)$ under $q_t$, we can use the formula 
$$
\E_{q_t}\{(\theta-\mu)^\top H_{ij}(\mu)(\theta-\mu)\} = \mathrm{tr}\{H_{ij}(\mu)\Sigma\}, 
$$
to get $\E_{q_t}\{f_{ij}(\theta)\} \approx f_{ij}(\mu)+\tfrac12 \mathrm{tr}\{H_{ij}(\mu)\Sigma\}$.  Then we plug this Taylor series-based approximation into the formula for evaluating $v^{(t)}$ in the Fisher update.  


\bibliography{reference}
\bibliographystyle{apalike}

\end{document}